\documentclass[10pt,twocolumn,letterpaper]{article}

\usepackage{cvpr}              

\usepackage{amsmath,amsfonts,bm}
\usepackage{xspace}
\usepackage{tabularx, booktabs, multirow, graphicx}
\usepackage{caption, subcaption}
\usepackage{colortbl}  
\usepackage{mathtools}
\makeatletter  
\newif\if@restonecol  
\makeatother

\usepackage[ruled,vlined,boxed,linesnumbered]{algorithm2e}
\usepackage{float}
\usepackage{placeins}
\usepackage[normalem]{ulem}  

\definecolor{cvprblue}{rgb}{0.21,0.49,0.74}
\usepackage[pagebackref,breaklinks,colorlinks,allcolors=cvprblue]{hyperref}

\def\paperID{40118} 
\def\confName{CVPR}
\def\confYear{2026}

\title{SparseCam4D: Spatio-Temporally Consistent 4D Reconstruction \\
from Sparse Cameras}

\author{
  Weihong Pan$^{1,2}$\footnotemark[1]\and 
  Xiaoyu Zhang$^{2}$\footnotemark[1]\and\
  Zhuang Zhang \and
  Zhichao Ye$^{2}$ \and 
  Nan Wang$^{2}$\and
  Haomin Liu$^{2}$\and
  Guofeng Zhang$^{1,2}$\footnotemark[2]\and
  \textnormal{\hspace{-1em}  $^1$State Key Lab of CAD\&CG, Zhejiang University \quad $^2$InSpatio Research }
  \\
} 

\begin{document}

\twocolumn[{%
\renewcommand\twocolumn[1][]{#1}%
\maketitle
\centering
\vspace{-0.5cm}
\includegraphics[width=0.99\linewidth]{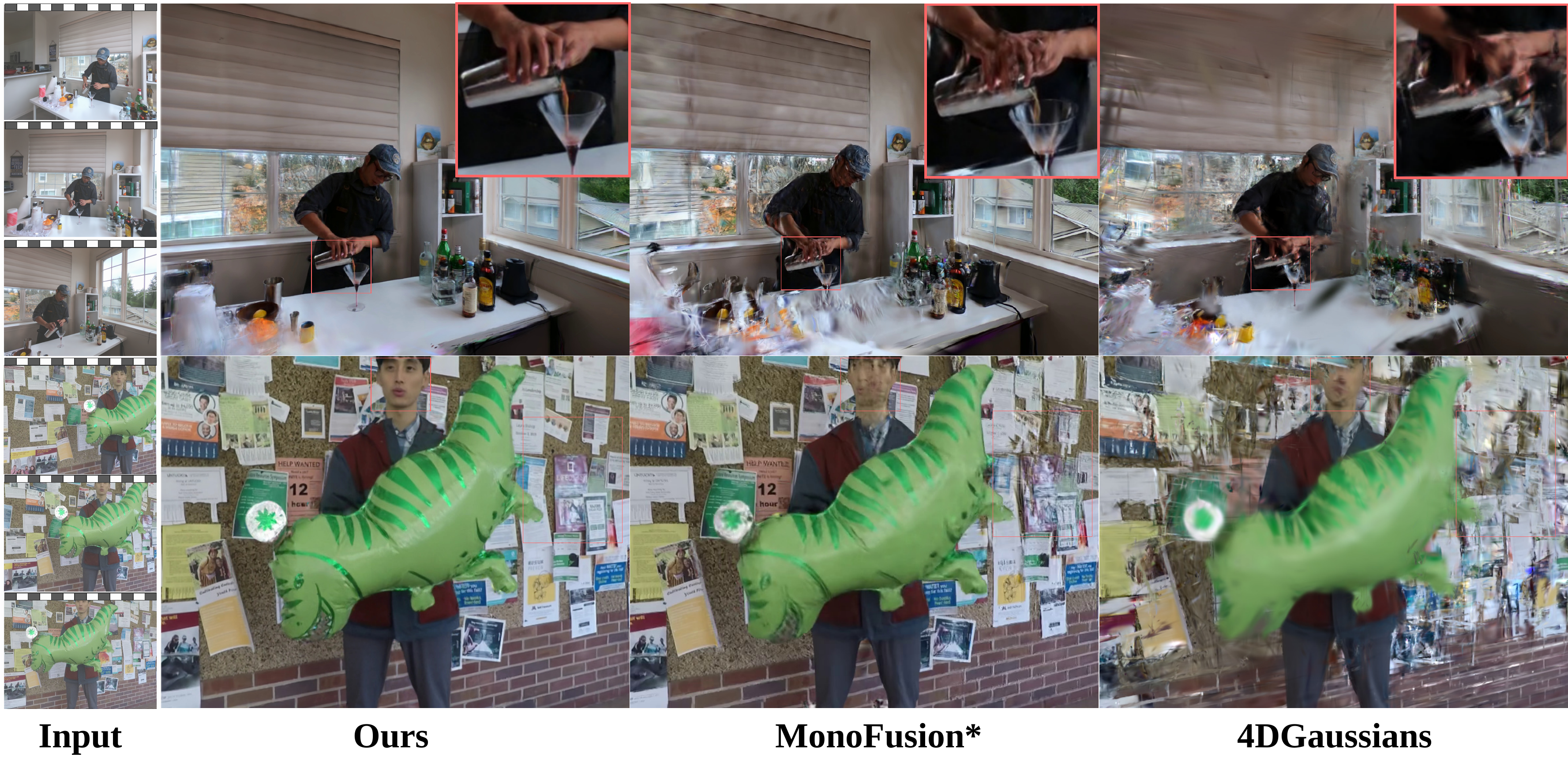}
\captionof{figure}{
\textbf{Novel view rendering comparison.} With as few as 2-3 cameras, our approach reconstructs high-quality dynamic scenes with spatio-temporal consistency and photorealistic quality. Please refer to our project page for additional dynamic results.}
\label{fig:teaser}
\vspace{5mm}
}]

\renewcommand{\thefootnote}{\fnsymbol{footnote}}
\footnotetext[1]{Equal contributions.} 
\footnotetext[2]{Corresponding authors.}

\begin{abstract}
High-quality 4D reconstruction enables photorealistic and immersive rendering of the dynamic real world. However, unlike static scenes that can be fully captured with a single camera, high-quality dynamic scenes typically require dense arrays of tens or even hundreds of synchronized cameras. Dependence on such costly lab setups severely limits practical scalability. To this end, we propose a sparse-camera dynamic reconstruction framework that exploits abundant yet inconsistent generative observations. Our key innovation is the Spatio-Temporal Distortion Field, which provides a unified mechanism for modeling inconsistencies in generative observations across both spatial and temporal dimensions. Building on this, we develop a complete pipeline that enables 4D reconstruction from sparse and uncalibrated camera inputs. We evaluate our method on multi-camera dynamic scene benchmarks, achieving spatio-temporally consistent high-fidelity renderings and significantly outperforming existing approaches.
\noindent Project page: \url{https://inspatio.github.io/sparse-cam4d/}
\end{abstract}  
\section{Introduction}
\label{sec:intro}
\begin{figure*}[t]
    \centering
    \begin{minipage}{0.99\linewidth} 
        \centering        
        \includegraphics[width=\linewidth]{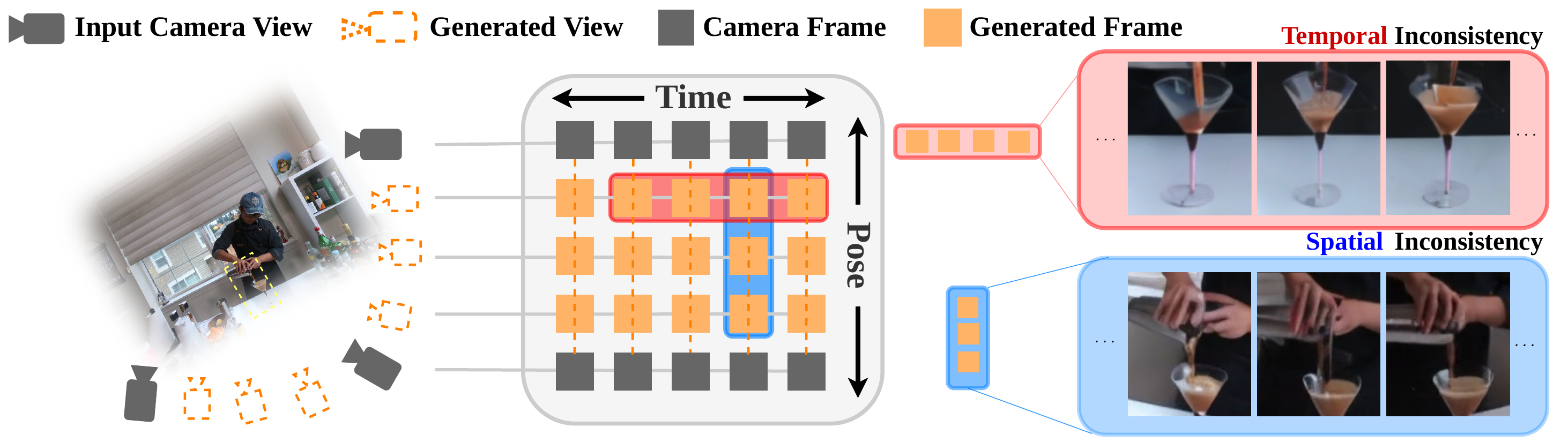}       
    \end{minipage}

    \caption{
    \textbf{Spatio-temporal inconsistency.} Real cameras (grey) capture consistent content of multi-view dynamic scene, while generative results (orange) include additional observations at different poses and time. Inconsistencies across poses at the same time are referred to as \textit{spatial inconsistencies}, and inconsistencies across time at the same pose are referred to as \textit{temporal inconsistencies}.
    }

    \label{fig:inconsistency}
    \vspace{-3mm}
\end{figure*}
Advances in dynamic scene novel view synthesis (NVS), particularly real-time 4D Gaussian Splatting (4DGS), have enabled high-fidelity dynamic rendering and hold great potential for applications in VR/AR, film production, short videos, live streaming, etc.~\citep{newcombe2015dynamicfusion,wu2024recent}.

However, immersive 4D reconstruction generally requires dense camera inputs. Unlike static scenes that can be captured with a single camera, high-quality dynamic scene benchmarks~\citep{li2022neural,sabater2017dataset,yoon2020novel} are typically constructed using dense camera arrays comprising approximately 20 synchronized cameras. Such costly lab setups severely hinder broader adoption and scalability.

Therefore, we aim to reconstruct high-fidelity dynamic scenes from sparse cameras.
Sparse-view 4D reconstruction remains largely unsolved~\citep{younis2025sparseview3dreconstructionrecent}, as spatio-temporally consistent reconstruction of complex motions relies heavily on dense observations. Sparse inputs aggravate the ill-posed nature of 4D reconstruction~\citep{jin2025diffuman4d}, making this task highly challenging.

Geometric regularization is an intuitive strategy for sparse-view reconstruction~\citep{younis2025sparseview3dreconstructionrecent}.
MonoFusion~\citep{wang2025monofusion}, followed with Shape-of-Motion~\citep{wang2024shapemotion4dreconstruction}, uses depth and tracking regularization to align 4D scene content and improve multi-view consistency in novel view synthesis.
However, these regularization techniques mainly focus on structural constraints and are insufficient to preserve accurate appearance, causing the rendering quality to quickly collapse under viewpoint shifts, as shown in Fig.~\ref{fig:teaser}, which prevents free-viewpoint exploration.

With the remarkable progress in Camera-Controlled Video Diffusion Models~\citep{yu2024viewcrafter,bai2025recammaster,bahmani2025ac3d}, another intuitive direction is to leverage such models to generate high-quality spatio-temporal data, thereby providing additional observation for 4D reconstruction. 
However, these photorealistic generated results often exhibit spatio-temporal inconsistencies, such as flickering surfaces and unstable object motions across views and time, as shown in Fig.~\ref{fig:inconsistency}, which undermine the coherence of dynamic scenes and cause severe blurring and artifacts.

To this end, our key innovation is the Spatio-Temporal Distortion Field (STDF), a lightweight mechanism enables unified modeling of inconsistencies in generative observations across both space and time. Notably, the STDF is discarded after training, thus introducing zero additional computational overhead to novel view rendering.
Moreover, due to the difficulty of obtaining accurate pose priors from sparse inputs, we conduct experiments using uncalibrated sparse views. Our pipeline jointly optimizes pose, rendering, and smoothness terms to produce spatio-temporally consistent dynamic reconstructions.

Finally,  we validate our approach on three standard 4D reconstruction benchmarks, including Neural 3D Video~\citep{li2022neural}, Technicolor~\citep{sabater2017dataset}, and Nvidia Dynamic Scenes~\citep{yoon2020novel}. To the best of our knowledge, this is the first work to achieve sparse-camera 4D reconstruction on dynamic scene benchmarks, evaluated across all camera views.

Our contributions are as follows: 
\textbf{(i)} We propose unified modeling of spatio-temporal inconsistencies in generative observations by introducing the Spatio-Temporal Distortion Field (STDF). 
\textbf{(ii)} We present a complete pipeline and optimization strategy that supports high-fidelity 4D scene reconstruction from uncalibrated sparse inputs. 
\textbf{(iii)} Through extensive experiments across multi-camera 4D benchmarks, our method demonstrates clear advantages over prior work, delivering photorealistic novel view renderings of dynamic scenes with spatio-temporal consistency under sparse-camera inputs.


\vspace{-2mm}
\section{Related Work}
\label{sec:related}
\subsection{Sparse-view dynamic reconstruction} 
Although numerous recent works~\cite{li2022neural, fridovich2023k, cao2023hexplane, wu20244d,yang2023real,yang2024deformable,lin2024gaussian,li2024spacetime,jing2025frnerf} reconstruct dynamic scenes from densely captured and well-synchronized multi-view video inputs. In contrast, sparse-view dynamic reconstruction is still in its early stages. Some studies have initially focused on reconstructing dynamic content of objects or human bodies using sparse cameras. Works such as~\cite{peng2021neural, weng2022humannerf, hu2024gauhuman, peng2021animatable} build a canonical static 3D space based on SMPL priors and learn a deformation field to map it to Gaussian primitives at different time steps.~\cite{jin2025diffuman4d} utilizes existing human datasets and train a spatio-temporal diffusion model under the guidance of SMPL priors to generate additional temporally consistent multi-view human videos for reconstruction. Research on sparse-view dynamic reconstruction in real-world scenes remains limited. A recent line of work explores dynamic reconstruction from monocular videos~\citep{wang2024shapemotion4dreconstruction,lei2024mosca,liu2025modgsdynamicgaussiansplatting,liang2025himor,park2025splinegs,wu20254d}. Nevertheless, these approaches are heavily dependent on monocular depth estimation and tracking-based regularization, and without multi-view constraints, the rendering of novel views quickly collapses under viewpoint shifts, making it difficult to obtain roamable, high-quality 4D scenes. MonoFusion~\citep{wang2025monofusion} improves upon this line by integrating monocular depth, 3D tracking information, and DINOv2~\citep{oquab2023dinov2} features, extending monocular 4D reconstruction methods to sparse-view settings. Nevertheless, results show that under these geometric regularizations, the rendering quality remains suboptimal, exhibiting floating artifacts, missing details, and geometric distortions, making it challenging to achieve photorealistic renderings in novel view.

\vspace{-1mm}
\subsection{Diffusion Model for Novel View Synthesis.} The rapid progress of video diffusion models (VDMs) has enabled generating continuous videos conditioned on a single image. Within the context of novel view synthesis, several works~\citep{liu2024sherpa3d,long2024wonder3d,shi2023mvdream,wu2024unique3d,ye2024dreamreward} directly focus on multi-view consistent video generation, but they are mostly constrained to object-centric settings. 
Scene-level video generation methods~\cite{sun2024dimensionx,wu2025cat4d,wang20254real} attempt to improve consistency via multi-view diffusion models. Yet, limited real-world multi-view video data and the inherent trade-off between spatial accuracy and generative capability lead to frequent inconsistencies, making precise 4D reconstruction in realistic environments highly challenging. Another line of work, camera-controlled video diffusion model, incorporates camera motion control during generation for controllable novel-view videos.~\cite{wang2024motionctrl} introduces a Camera Motion Control Module that injects camera extrinsic parameters into temporal transformers.~\cite{xu2024camco,he2024cameractrl,bai2025recammaster} further improve controllability by replacing explicit camera pose parameters with Plücker ray embeddings.~\cite{yu2024viewcrafter} incorporate point cloud priors to improve motion control stability. However, generated frames remain spatially and cross-view inconsistent, limiting their suitability in 4D reconstruction. We address this by enforcing consistency across space, time, and viewpoints.

\vspace{-2mm}
\section{Method}
\label{sec:method}
\begin{figure*}[t]
    \centering
    \begin{minipage}{0.99\linewidth} 
        \centering        
        \includegraphics[width=\linewidth]{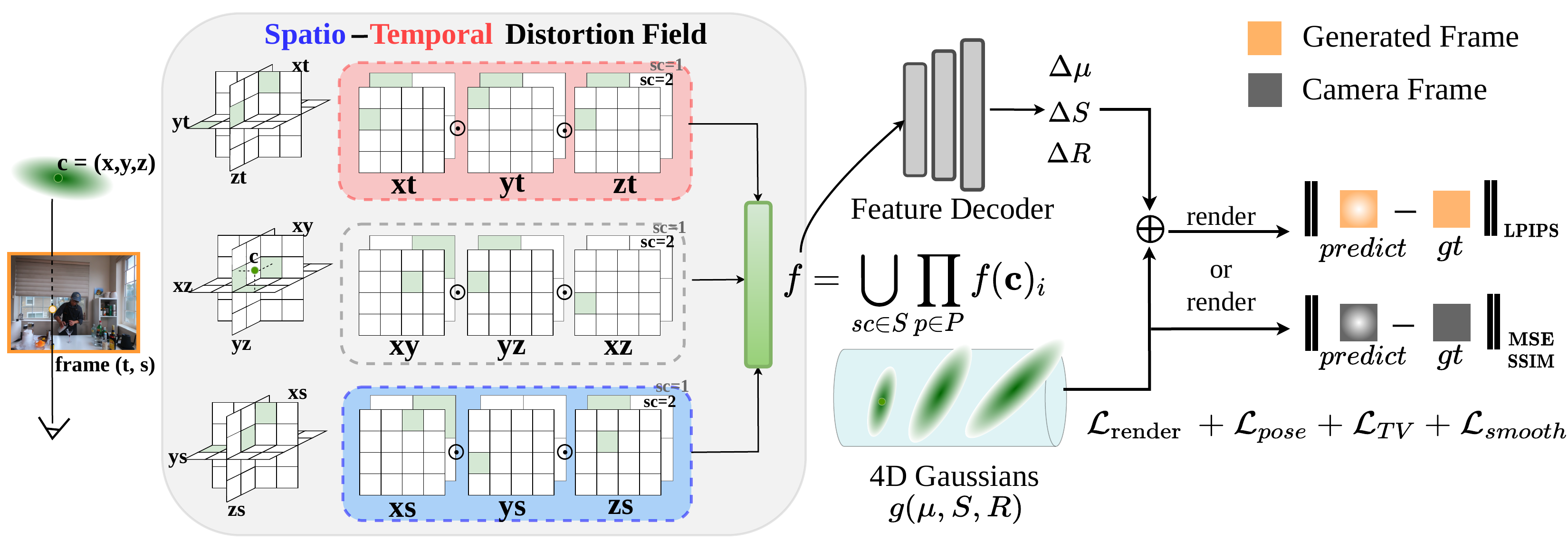}       
    \end{minipage}
    \caption{ 
        \textbf{Method overview.} Given a generated frame at temporal index $t$ and pose index $s$, each 4D Gaussian at $c=(x,y,z)$ is projected onto the planes of the Spatio-Temporal Distortion Field to obtain deformation features, which are then decoded by a small MLP to produce the deformation values. 
        We use separate photometric losses for real and generated frames, and additionally introduce regularization terms on pose, feature plane, and spatial smoothness to enhance optimization stability.
    }
    \label{fig:framework}
    \vspace{-3mm}
\end{figure*}
In this section, we introduce a novel framework that harnesses video diffusion models as a source of auxiliary visual observations, facilitating 4D scene reconstruction when only a limited set of input views is available. Sec~\ref{subsec:framework} introduces the overall framework. Sec~\ref{subsec:field} details the core component - Spatio-Temporal Distortion Field - which helps to incorporate generated images into 4D scene reconstruction. Finally, Sec~\ref{subsec:optimization} describes the optimization scheme.

\subsection{Preliminary}
\label{subsec:preliminary}
\textbf{4D Gaussian Splatting.} A line of 4DGS approaches directly models Gaussian primitives in 4D space to represent the dynamic scene. In this paradigm, the temporal axis is treated as an additional independent coordinate dimension, such that 3D Gaussians are directly lifted into 4D. Each 4D Gaussian is parameterized by its center position $\boldsymbol{\mu}=(\mu_x,\mu_y,\mu_z,\mu_t)$ and a covariance matrix $\boldsymbol{\Sigma}\in\mathbb{R}^{4\times4}$, where $\boldsymbol{\Sigma}$ is decomposed into a scaling matrix $\boldsymbol{S}=\text{diag}(s_x,s_y,s_z,s_t)$ and a rotation matrix $\boldsymbol{R}\in\mathbb{R}^{4 \times 4}$. Same as 3D Gaussians, each 4D primitive maintains a set of SH coefficients and an opacity $\alpha$.

To parameterize 4D rotation matrix in Euclidean space, algebraic geometry tools such as pair of isotropic rotations~\cite{yang2023real} $\boldsymbol{R}=L(\boldsymbol{q}_l)R(\boldsymbol{q}_r)$ or a normalized 4D rotor \textbf{r} with 8 coefficients~\cite{duan20244d} $\boldsymbol{R}=\mathcal{F}_{map}(\mathcal{F}_{norm}(\textbf{r}))$ are employed, which are mathematically equivalent. At a given timestamp $t$, temporal slicing is performed to project the attributes of each 4D Gaussian into the corresponding 3D subspace as follows~\cite{duan20244d}:
\vspace{-1mm}
\begin{equation}
\begin{split}
    \mathcal{G}_{3D}(\boldsymbol{x},t)&=
        \text{e}^{-\frac{1}{2}\lambda(t-\mu_t)^2}
        \text{e}^{-\frac{1}{2}[\boldsymbol{x}-\boldsymbol{\mu}(t)]^T\boldsymbol{\Sigma}_{3D}^{-1}[\boldsymbol{x}-\boldsymbol{\mu}(t)]}, \\
\end{split}
\end{equation}

\vspace{-1mm}
\noindent The rendering process follows the standard differential splatting procedure of 3DGS, while the densification is performed in both spatial and temporal dimensions.

\noindent{\textbf{K-planes Factorization.}} K-planes~\citep{fridovich2023k} introduces a simple and interpretable representation for arbitrary $d$-dimensional scenes, referred to as \textit{K-planes factorization}. In this framework, $k=\binom{d}{2}$ planes are employed to represent every combinations of two dimensions. Taking an $d$-dimensional coorinate as input, K-planes maps it to a feature vector, which is then decoded by a tiny MLP to obtain target attribute value. For dynamic 4D scenes, this results in the so-called \textit{hex-planes}, consisting of three space-only planes $xy$, $xz$, $yz$ and three space-time planes $xt$, $yt$, $zt$. Such a representation has been widely adopted in both deformation-based NeRF and 4DGS methods. 
NeRFs use it to estimate a world-to-canonical mapping~\citep{fridovich2023k} $\mathcal{M}:(\boldsymbol{p},t)\rightarrow\Delta\boldsymbol{p}$, where $\boldsymbol{p}$ reveals to the world spatial point and $t$ is the target time. Then the vanilla NeRF pipeline is applied with canonical spatial point $\boldsymbol{p}+\Delta\boldsymbol{p}$ and view direction $\boldsymbol{d}$ as input.
4DGS using it to compute the canonical-to-world mapping~\citep{wu20244d} $\mathcal{F}:(\mathcal{G},t)\rightarrow\Delta\mathcal{G}$ for each attribute of a canonical 3D Gaussian primitive $\mathcal{G}$ at time $t$. An image $I$ with view matrix $\textbf{M}=[\textbf{R}|\textbf{T}]$ is rendered by the differential splatting with the deformed 3D Gaussians $\mathcal{G}'$ following $I=\mathcal{S}(\textbf{M}, \mathcal{G}')$, where $\mathcal{G}'=\mathcal{G}+\Delta\mathcal{G}$. 

\subsection{Framework}
\label{subsec:framework}
Given $N$ sparse input camera videos with $L$ frames, our goal is to optimize a 4DGS model using auxiliary generated sequences. For simplicity, we refer to the set of images from $N$ input video sequences and their corresponding camera poses as \textit{input views} $V_{I}=\{(I_s^t,[\textbf{R}|\textbf{T}]_s)|t=0,...,L;s=0,...,N\}$, and the set of images $I$ from $M$ generated video sequences with their poses $[\textbf{R}|\textbf{T}]$ as \textit{generated views} $V_{G}=\{(I_s^t,[\textbf{R}|\textbf{T}]_s)|t=0,...,L;s=0,...,M\}$. The 4DGS model is trained with $V_I+V_G$.

The noising and denoising process of video diffusion models introduces severe geometric inconsistencies across space and time during generation. If such generated views are directly used for scene reconstruction, these inconsistencies will significantly degrade the geometric consistency of 4DGS, leading to noticeable artifacts. Therefore, when leveraging such diffusion-based observations to assist reconstruction, it is crucial to extract and disentangle these inconsistencies to construct canonical 4D Gaussians.

Our framework consists of 4D Gaussians $\mathcal{G}_{4D}$ and a spatio-temporal distortion field $\mathcal{F}$ that models the inconsistencies in each generated view $V_G^{t,s}\in V_G$, formally represented as:
\begin{equation}
    \mathcal{F}:(\mathcal{G}_{4D},t,s)\rightarrow \Delta\mathcal{G}_{4D}.
\end{equation}
where $t$ denotes the time index, while $s$ denotes the pose index, as illustrated in Fig.~\ref{fig:inconsistency}. Through the proposed distortion field, the variation of canonical 4D Gaussians on a generated view can be obtained, which then yields the distorted 4D Gaussians $\mathcal{G}_{4D}'=\mathcal{G}_{4D}+\Delta\mathcal{G}_{4D}$. For a generated view indexed by $(t,s)$, our framework converts the canonical 4D Gaussians to a set of distorted Gaussians while preserving compatibility with differentiable splatting.

\subsection{Spatio-Temporal Distortion Field}
\label{subsec:field}
Specifically, the spatio-temporal distortion field $\mathcal{F}$ consists of an Ennea-plane representation and a lightweight multi-head MLP serving as the fused feature decoder. We factorize the 5D volume defined by $(x,y,z,t,s)$ into $k=\binom{5}{2}=10$ two-dimensional planes, each corresponding to a pair of dimensions. Since the combination $(t,s)$ does not encode any form of distortion, this plane is omitted. Consequently, such factorization decomposes the 5D neural voxel into nine multi-resolution 2D feature planes $\boldsymbol{P} = \{\boldsymbol{P}_{xy},\boldsymbol{P}_{xz},\boldsymbol{P}_{yz},\boldsymbol{P}_{xt},\boldsymbol{P}_{yt},\boldsymbol{P}_{zt}$,$\boldsymbol{P}_{xs},\boldsymbol{P}_{ys},\boldsymbol{P}_{zs}\}$. Each feature plane is defined as $\boldsymbol{P}_{ij}\in\mathbb{R}^{lN_i\times lN_j\times h}$, where $h$ denotes the feature dimension, $N_i$ and $N_j$ represent the basis resolution of the corresponding two axes, and $l$ is the scale factor for multi-resolution structure.

Given a 5D coordinate $\boldsymbol{c}=\{x,y,z,t,s\}$, the corresponding feature vector is obtained as follows. First, each dimension of $\boldsymbol{c}$ is normalized to its resolution range $[0,N_i)$, and then coordinate $\boldsymbol{c}$ is projected onto the nine planes aforementioned. The feature of $\boldsymbol{c}$ on each plane is extracted via bilinear interpolation, formally:
\begin{equation}
\begin{split}
    \boldsymbol{f}(\boldsymbol{c})_c&=\text{interp}(\boldsymbol{P}_c,\pi_c(\boldsymbol{c})),\\ 
    &c\in\{xy,xz,yz,xt,yt,zt,xs,ys,zs\}
\end{split}
\end{equation}
where $\pi_c$ denotes the projection of $\boldsymbol{c}$ onto the corresponding plane, and `$\text{interp}$' indicates the bilinear interpolation over the 2D grid. The features extracted from the feature planes are then fused by element-wise multiplication to obtain an $h$-dimensional feature vector at a given resolution. Features across different resolutions are then concatenated to form the final features.
\begin{equation}
    \boldsymbol{f}(\boldsymbol{c}) = \bigcup_{sc}\prod_{\boldsymbol{P}_c\in\boldsymbol{P}}\boldsymbol{f}(\boldsymbol{c})_c
\end{equation}
These features are decoded by a multi-head MLP decoder $\mathcal{D}=\{\phi, \phi_{\boldsymbol{p}},\phi_{\boldsymbol{q}_l},\phi_{\boldsymbol{q}_r},\phi_{\boldsymbol{s}}\}$ into the distortion of various 4D Gaussian attributes, including 
position $\Delta\boldsymbol{\mu}=\phi_{\boldsymbol{\mu}}(\phi(\boldsymbol{f}))$, 
rotation $\Delta\boldsymbol{q}_l=\phi_{\boldsymbol{q}_l}(\phi(\boldsymbol{f})),\Delta\boldsymbol{q}_r=\phi_{\boldsymbol{q}_r}(\phi(\boldsymbol{f}))$, 
and scaling $\Delta\boldsymbol{s}=\phi_{\boldsymbol{s}}(\phi(\boldsymbol{f}))$. 
Then, the distorted attributes can be computed as:
\vspace{-2mm}
\begin{equation}    
\begin{aligned}
(\boldsymbol{\mu}',\boldsymbol{q}_l',\boldsymbol{q}_r',\boldsymbol{s}')=(\boldsymbol{\mu}+\Delta\boldsymbol{\mu},\boldsymbol{q}_l+\Delta\boldsymbol{q}_l,\\
\boldsymbol{q}_r+\Delta\boldsymbol{q}_r,\boldsymbol{s}+\Delta\boldsymbol{s})
\end{aligned}
\end{equation}

\vspace{-2mm}
During training, the distorted Gaussians are used to render the generated views, while the original Gaussians are used to render the real views. After training, the distortion terms are discarded, and only the canonical 4D Gaussian is retained.

\subsection{Optimization}
\label{subsec:optimization}
\textbf{Pose Optimization.} Because of the nconsistencies present in the generated video frames, the alignment accuracy is compromised when using traditional COLMAP~\citep{schoenberger2016sfm,schoenberger2016mvs} for estimating camera extrinsics. To mitigate this issue, we propose simultaneous optimization of camera extrinsics along with the 4D Gaussian attributes, treating camera extrinsics as learnable variables likewise.

\noindent{\textbf{Loss Function.}} For input views, we apply the standard photometric loss. The photometric loss includes an $\mathcal{L}_1$ RGB loss and a D-SSIM loss~\citep{wang2004image}.

\vspace{-3mm}
\begin{equation}
    \mathcal{L}_{\text{input}}=(1-\lambda)\mathcal{L}_1+\lambda\mathcal{L}_{\text{D-SSIM}}.
\end{equation}

For generated views, applying standard photometric loss directly leads to degraded reconstruction quality due to the inherent distortions in generated frames. To address this, we incorporate perceptual loss~\citep{zhang2018unreasonable} to supervise texture and reconstruction similarity.
\vspace{-1mm}
\begin{equation}
    \mathcal{L}_{\text{gen}}=\lambda_1\mathcal{L}_1+\lambda_2\mathcal{L}_{lpips}.
\end{equation}

In order to prevent the optimized pose from deviating significantly from their original initialization, a regularization term is introduced:
\begin{equation}
    \mathcal{L}_{\text{pose}}=\lambda_p(||\boldsymbol{T}-\hat{\boldsymbol{T}}||+||\boldsymbol{q}-\hat{\boldsymbol{q}}||),
\end{equation}
where $\boldsymbol{T}$ and $\boldsymbol{q}$ represent the optimized translation and rotation of a camera, $\hat{\boldsymbol{T}}$ and $\hat{\boldsymbol{q}}$ are the corresponding initial extrinsics obtained from COLMAP~\citep{schoenberger2016sfm,schoenberger2016mvs}, and parameter $\lambda_{\text{p}}$ balances the camera optimization term with other loss components.

Additionally, following K-Planes~\citep{fridovich2023k}, a grid-based total variation loss $\mathcal{L}_{\text{TV}}$ is also applied for spatial smoothness. Since the distortions in the generated images from~\citet{yu2024viewcrafter} are continuous along the pose axis but exhibit abrupt changes along the time axis, we apply a smoothness regularization over pose axis with a second derivative filter:
\begin{equation}
\begin{aligned}
\mathcal{L}_{\text{smooth}}
&= \lambda_s 
   \frac{1}{|C|}
   \sum_{c\in C} 
   \frac{1}{N_i N_s}
   \sum_{i,s}
   \bigl\| (\boldsymbol{P}_c^{i,s-1}-\boldsymbol{P}_c^{i,s}) \\
&\quad -(\boldsymbol{P}_c^{i,s}-\boldsymbol{P}_c^{i,s+1})\bigr\|_2^2, \\
C&=\{x_s, y_s, z_s\}.
\end{aligned}
\end{equation}
where $i$, $s$ are indices on plane $\boldsymbol{P}_c$, $N_i$ and $N_s$ represent the resolution of each axis.
Overall, the total loss can be formulated as:
\begin{equation}
    \mathcal{L} = \mathcal{L}_{\text{input}}+\mathcal{L}_{\text{gen}}+\mathcal{L}_{\text{pose}}+\mathcal{L}_{\text{TV}}+\mathcal{L}_{\text{smooth}}
\end{equation}
\section{Experiments}
\label{sec:experiment}
\subsection{Experimental Setups}
\label{subsec:expSetup}
\subsubsection{Datasets.}
\label{subsubsec:dataset}
We conduct extensive experiments on three real-world datasets: Neural 3D Video, Technicolor, and Nvidia Dynamic Scenes. Training is performed on two or three selected views that adequately cover the scene content, and evaluation is carried out on \textbf{all the remaining views}. For more experimental setting details, please refer to the supplementary material.

\noindent{\textbf{Neural 3D Video Dataset.}}~\citep{li2022neural} This dataset comprises six indoor multi-view video sequences, recorded simultaneously by 18-21 synchronized cameras at a resolution of 2704$\times$2028 and 30 fps. Following~\cite{yang2023real}, we perform both training and evaluation on downsampled videos by a factor of two, using 300 frames per scene.

\noindent{\textbf{Technicolor Dataset.}}~\citep{sabater2017dataset} This dataset includes five indoor multi-video sequences acquired with a 4$\times$4 synchronized camera array at a resolution of 2048$\times$1088. Following~\cite{attal2023hyperreel}, we conduct both training and evaluation at full resolution, using 50 frames per scene.

\noindent{\textbf{Nvidia Dynamic Scenes Dataset.}}~\citep{yoon2020novel} This dataset contains of six outdoor multi-view video sequences, each captured by 12 synchronized cameras at 1920$\times$1080 and 60Hz. We use half-resolution frames and 100 frames per scene for training and evaluation.

\newcommand\rowrule{\rule{0pt}{10pt}}

\begin{table*}[t]    
    \caption{\textbf{Qualitative comparisons on Technicolor~\citep{sabater2017dataset}, Neural 3D Video~\citep{li2022neural}, and Nvidia Dynamic Scenes~\citep{yoon2020novel} Datasets.} 
    The \colorbox{red!30}{first} and \colorbox{yellow!30}{second}best performances are highlighted in red and yellow. Our method shows superior performance compared to all baseline methods across all metrics. Note that MonoFusion$^*$ is our reproduced version.}
    \vspace{-1mm}
    \centering
    \resizebox{0.95\linewidth}{!}{%
    \begin{tabular}[t]{l|ccc|ccc|ccc}    
        \toprule[2pt]
        \multirow{2}{*}{Method}
        &\multicolumn{3}{c|}{Technicolor}
        &\multicolumn{3}{c|}{Neural 3D Video}
        &\multicolumn{3}{c}{Nvidia Dynamic Scenes}\\
        &\text{PSNR}$\uparrow$&\text{SSIM}$\uparrow$&\text{LPIPS}$\downarrow$
        &\text{PSNR}$\uparrow$&\text{SSIM}$\uparrow$&\text{LPIPS}$\downarrow$
        &\text{PSNR}$\uparrow$&\text{SSIM}$\uparrow$&\text{LPIPS}$\downarrow$\\
        \hline    
        \rowrule HyperReel~\citep{attal2023hyperreel}      
        &14.14&0.453&0.616
        &15.63&0.582&0.500
        &19.88&0.528&0.396\\
        \rowrule 4DGaussians~\citep{wu20244d}  
        &16.20&0.505&0.552
        &17.40&0.673&0.320
        &16.81&0.372&0.516\\
        \rowrule 4D-Rotor~\citep{duan20244d} 
        &14.85&0.426&0.581
        &18.20&0.708&0.357
        &19.38&0.508&0.389\\
        \rowrule RealTime4DGS~\citep{yang2023real} 
        &16.53&0.510&0.542
        &17.31&0.649&0.442        
        &17.91&0.479&0.426\\
        \rowrule MonoFusion*~\citep{wang2025monofusion} 
        &\colorbox{yellow!30}{17.97}&\colorbox{yellow!30}{0.578}&\colorbox{yellow!30}{0.352}
        &\colorbox{yellow!30}{18.43}&\colorbox{yellow!30}{0.738}&\colorbox{yellow!30}{0.270}
        &\colorbox{yellow!30}{20.22}&\colorbox{yellow!30}{0.590}&\colorbox{yellow!30}{0.192}\\
        \hline
        \rowrule Ours 
        &\colorbox{red!30}{23.15}&\colorbox{red!30}{0.728}&\colorbox{red!30}{0.299} 
        &\colorbox{red!30}{21.91}&\colorbox{red!30}{0.789}&\colorbox{red!30}{0.258}
        &\colorbox{red!30}{24.81}&\colorbox{red!30}{0.794}&\colorbox{red!30}{0.150}\\
        \bottomrule[2pt]
    \end{tabular}
    }
\label{tab:exp}
\end{table*}
\begin{figure*}[t]
    \centering
    \begin{minipage}{0.98\linewidth} 
        \centering        
        \includegraphics[width=\linewidth]{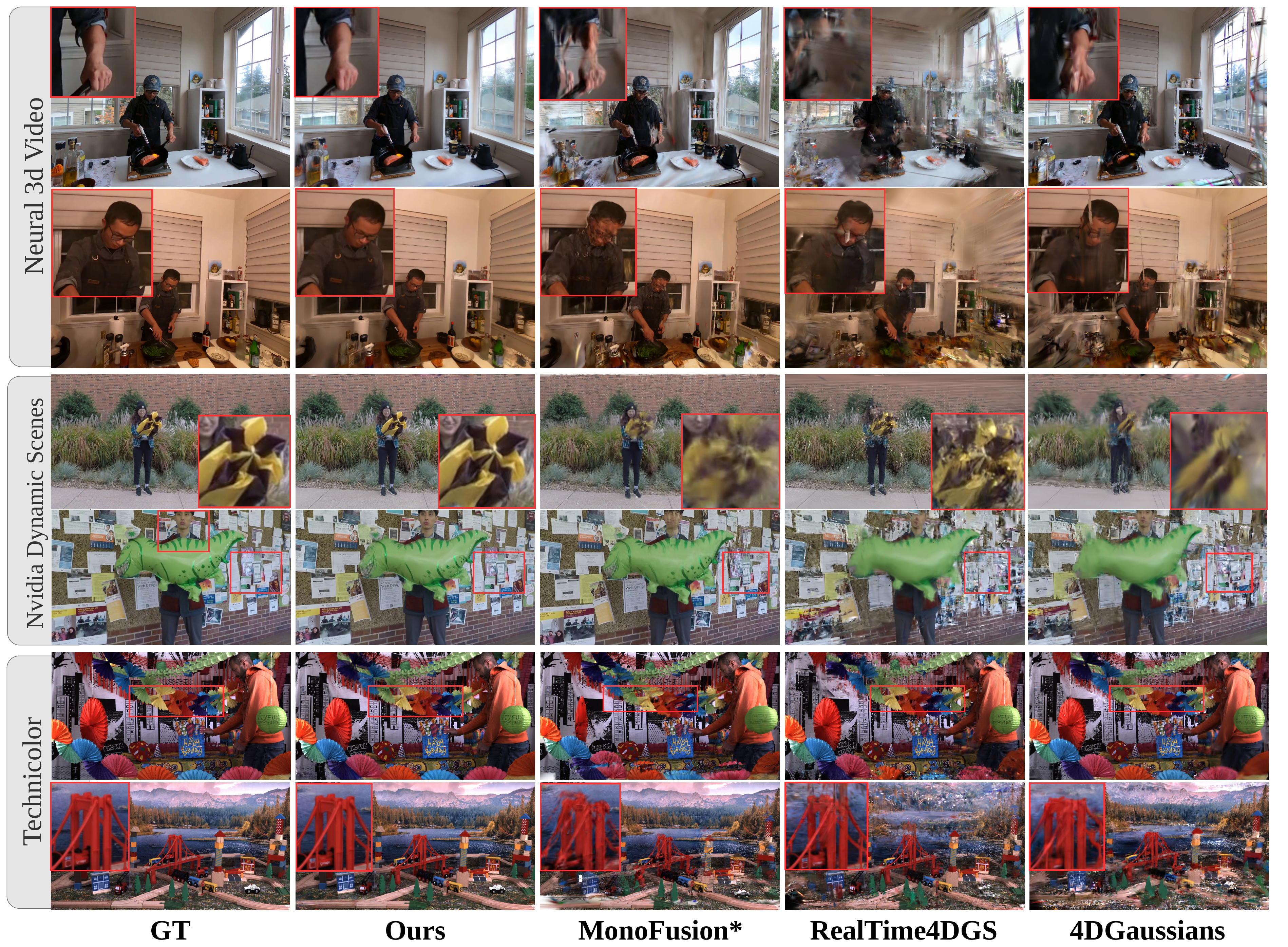}       
    \end{minipage}
    \vspace{-2mm}
    \caption{ 
        \textbf{Qualitative Comparisons of different methods on Technicolor~\citep{sabater2017dataset}, Neural 3D Video~\citep{li2022neural}, and Nvidia Dynamic Scenes~\citep{yoon2020novel} Datasets}. We conduct comparisons with representative dynamic scene reconstruction methods: MonoFusion~\citep{wang2025monofusion}, 4DGS~\citep{wu20244d}, 4D-Rotor~\citep{duan20244d}, and Realtime4DGS~\citep{yang2023real}. MonoFusion$^*$ is our reproduced version. Our method significantly outperforms other baselines, producing visually reliable results with sharper details. Please zoom in for more details. Additional qualitative comparisons are included in the supplementary material.
    }
    \label{fig:exp} 
    \vspace{-3mm}
\end{figure*}
\vspace{5mm}

\begin{table}[htbp]    
    \caption{\textbf{Ablation studies on STDF.} We randomly select one representative scene from Technicolor~\citep{sabater2017dataset} and Nvidia Dynamic Scenes~\citep{yoon2020novel} to ablate  Spatio-Temporal Distortion Field.}
    \centering
    \resizebox{0.95\linewidth}{!}{%
    \begin{tabular}[t]{l|cc|cc}    
        \toprule[2pt]
        \multirow{2}{*}{Setting}        
        &\multicolumn{2}{c|}{\textbf{Train}}
        &\multicolumn{2}{c}{\textbf{Jumping}}
        \\
        &\text{LPIPS}$\downarrow$&\text{SSIM}$\uparrow$
        &\text{LPIPS}$\downarrow$&\text{SSIM}$\uparrow$\\
        \hline    
        \rowrule w/o distortion field     &0.608&0.426 &0.319&0.674\\                                          
        \rowrule w/o time axis            &0.458&0.480 &0.279&0.738\\                                          
        \rowrule w/o pose axis            &0.469&0.462 &0.268&0.733\\
        \hline
        \rowcolor{gray!30}\rowrule Ours         
                                        &\textbf{0.264}&\textbf{0.656}
                                        &\textbf{0.170}&\textbf{0.793} \\                                        
        \bottomrule[2pt]
    \end{tabular}
    }
\label{tab:ablation}
\end{table}
\begin{figure}[t]
  \centering
   \includegraphics[width=\linewidth]{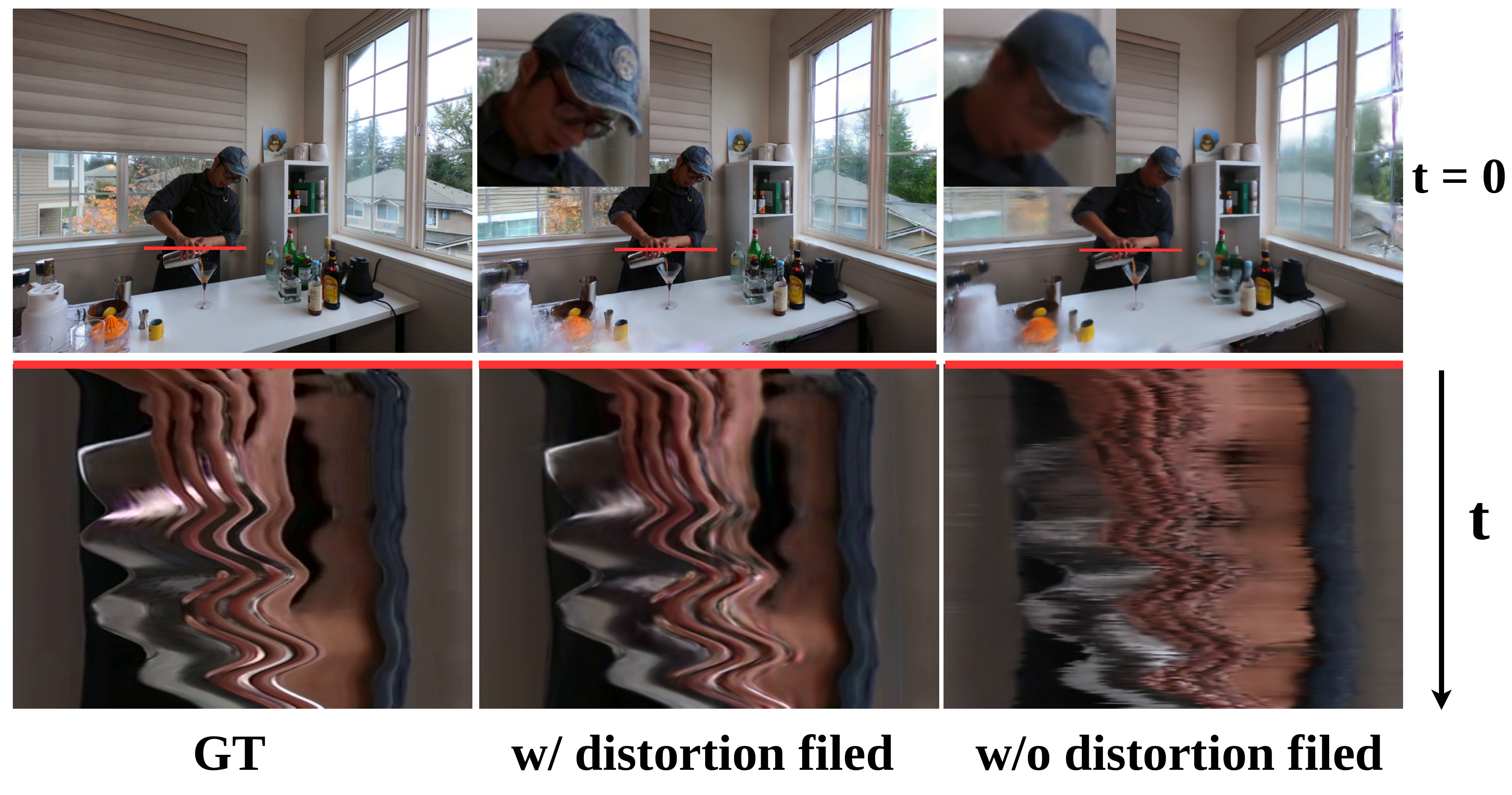}
   \vspace{-7mm}
\caption{\textbf{Spatio-Temporal Consistency.} Rendering results (top) and space-time slices (bottom) constructed by concatenating the red pixel locations across all time steps, demonstrate that direct reconstruction from diffusion observations leads to severe blur and temporal instability(e.g., the moving hand at the bottom right).}
   \label{fig:slice}
\end{figure}

\vspace{-3mm}
\subsubsection{Implementation Details.}
\label{subsubsec:imple}
Our framework is implemented with Pytorch~\citep{paszke2019pytorch} and optimized for 30,000 iterations per scene. At each iteration, one input-view image and one generated-view image are randomly sampled. The first 3000 iterations are trained with a vanilla setting as a warm-up stage, followed by training with both camera optimization and the distortion field, where camera optimization is stopped after 7000 iterations. For generated views, we employ a camera-controlled image-to-video diffusion model~\cite{yu2024viewcrafter} which generates $L=25$ frames per sequence. 
For uncalibrated generated views and cameras, we obtain coarse pose and point cloud initialization using COLMAP~\citep{schoenberger2016sfm,schoenberger2016mvs} at $t=0$.
The weighting factors $\lambda$, $\lambda_1$, $\lambda_2$, $\lambda_p$ and $\lambda_s$ are set to 0.2, 0.02, 0.2, 0.1 and 10$^{-4}$, respectively. All experiments are conducted on a single Nvidia A800 GPU to ensure fair comparisons.

\noindent{\textbf{Aligning camera poses on test views.}} In standard 4DGS reconstruction process, the exact camera poses of the test views are usually available, as they can be estimated together with the training views under a shared coordinate system before reconstruction. However, in our experimental setting, the poses of test views are unknown and pose optimization is applied during training. Therefore, aligning test-view poses before rendering is essential. Following the protocol of~\cite{fan2024instantsplat}, the trained 4DGS is kept fixed while only the test-view camera poses are refined. This optimization minimizes the $l_1$ photometric loss between rendered and ground-truth images, yielding more accurate alignment of the rendered results with test views. Such alignment eliminates errors caused by inaccurate poses, ensuring a fairer comparison.

\subsection{Results of Dynamic Novel View Synthesis}
\label{subsec:comparisons}
We adopt PSNR, SSIM~\citep{wang2004image}, and LPIPS~\citep{zhang2018unreasonable} as evaluation metrics for comparing the rendering quality of our method against baselines.
As the compared baselines do not include pose optimization, we adopt the ground-truth poses for both training and rendering. While this setup provides the baselines with a mild advantage, it does not overcome the fundamental challenge caused by sparse camera inputs. Qualitative and quantitative results are shown in Fig.~\ref{fig:exp} and Tab.~\ref{tab:exp}, respectively. The visualization results and metrics demonstrate that our approach consistently produces sharper details, more stable dynamics, and fewer artifacts under sparse-camera settings, achieving substantially better visual quality and spatio-temporal consistency than all baseline methods. 
For general 4DGS methods such as 4DGaussians~\citep{wu20244d} and RealTime4DGS~\citep{yang2023real}, their performance drops significantly due to the ill-posed nature of sparse-camera condition, producing broken geometry, noisy renderings and missing details in dynamic regions. These results highlight their reliance on dense and well-aligned inputs.
Compared with general 4DGS, MonoFusion~\citep{wang2025monofusion} benefits from various geometric priors and thus offers relatively high-quality initialization. This leads to noticeable improvements in static background regions. However, it still produces artifacts and geometric misalignments, especially in complex dynamic regions. For instance, in the Nvidia Dynamic Scenes dataset, rapid motions and occlusions cause its priors to fail to capture fine-grained dynamics, resulting in unreliable constraints and degraded reconstructions. The limitation mainly stems from the insufficient quality and robustness of the imposed priors under highly dynamic settings.
In contrast, our method leverages generative priors and explicitly disentangles distortions using the proposed spatio-temporal distortion field, achieving spatio-temporal consistency even under challenging conditions, including large view ranges and complex foreground dynamics. For example, in the Neural 3D Video dataset, our method reconstructs both static and dynamic regions with fine-grained details, while in the Nvidia Dynamic Scenes dataset, it preserves temporal stability despite outdoor motions. These results demonstrate that our method not only mitigates artifacts and geometric failures observed in baselines but also provides robust and reliable reconstructions across diverse datasets.

\subsection{Ablation and Analysis}
\label{subsec:ablation}
LPIPS and SSIM are adopted for ablation, as these two metrics better capture structural similarity and detail fidelity and reflect the impact of each component more accurately. Further details are provided in the supplementary material.

\noindent{\textbf{Effectiveness of the Spatio-Temporal Distortion Field.}} The `\textit{w/o distortion field}' variant removes the proposed Spatio-Temporal Distortion Field and directly reconstructs 4D scenes with generated images. As shown in Tab.~\ref{tab:ablation} and Fig.~\ref{fig:slice}, it produces severe blur result due to the spatio-temporal inconsistencies introduced by generative observations, whereas our distortion field substantially mitigates such problem and significantly improves rendering quality. The space-time slice depicted in Fig.~\ref{fig:slice} (bottom) is obtained by concatenating the red pixels on Fig.~\ref{fig:slice} (top) across all time steps. Compared with reconstructing directly with generated images, our method yields more temporally coherent outputs that closely match the ground-truth motion, validating the effectiveness of the proposed distortion-aware 4DGS design.

In addition, to further validate the effectiveness of STDF for jointly capturing spatial and temporal inconsistencies, we conducted the following two experiments. The `\textit{w/o temporal index}' and `\textit{w/o pose index}' variants remove the $t$-axis and $s$-axis in the STDF respectively, thereby modeling inconsistencies only along spatial or temporal dimensions. As shown in Tab.~\ref{tab:ablation}, the noticeable performance degradation indicates that generative inconsistencies manifest across both space and time. This validates the necessity of our two-dimensional temporal design, which more effectively aligns spatio-temporal content and improves rendering quality.

\noindent{\textbf{Effectiveness of Pose Optimization.}} In the \textit{w/o pose optimization} variant, camera poses are fixed during training without refinement. As reported in Tab.~\ref{tab:ablation-loss}, incorporating pose optimization substantially enhances reconstruction quality. This demonstrates that generative distortions can severely bias pose estimation, and correcting them during training is necessary for consistent 4D reconstruction.

\noindent{\textbf{Ablation on Spatio-Temporal Regularization.}} 
We employ $\mathcal{L}_{tv}$ on spatial planes and $\mathcal{L}_{smooth}$ over pose axis to enforce smoothness constraints. Results in Tab.~\ref{tab:ablation-loss} demonstrate that both terms are necessary for reconstructing high-quality 4D scenes. 
They reflect the continuity in spatio-temporal distortions, consistent with the gradual deformations observed in diffusion-generated outputs. 

\begin{table}[t]    
    \caption{\textbf{Ablation studies on pose optimization and loss components.} We randomly select one representative scene from Technicolor~\citep{sabater2017dataset} and Nvidia Dynamic Scenes~\citep{yoon2020novel} Datasets to ablate pose optimization and loss components.}
    \centering
    \resizebox{0.95\linewidth}{!}{%
    \begin{tabular}[t]{l|cc|cc}    
        \toprule[2pt]
        \multirow{2}{*}{Setting}        
        &\multicolumn{2}{c|}{\textbf{Train}}
        &\multicolumn{2}{c}{\textbf{Jumping}}
        \\
        &\text{LPIPS}$\downarrow$&\text{SSIM}$\uparrow$
        &\text{LPIPS}$\downarrow$&\text{SSIM}$\uparrow$\\
        \hline
        \rowrule w/o pose optimization      &0.336&0.569 &0.217&0.754\\             
        \rowrule w/o $\mathcal{L}_{lpips}$  &0.285&0.646 &0.185&0.782\\
        \rowrule w/o $\mathcal{L}_{TV}$     &0.309&0.616 &0.191&0.768\\
        \rowrule w/o $\mathcal{L}_{smooth}$ &0.308&0.619 &0.191&0.770\\
        \hline
        \rowcolor{gray!30}\rowrule Ours                                                 &\textbf{0.264}&\textbf{0.656}&\textbf{0.170}&\textbf{0.793} \\                                     
        \bottomrule[2pt]
    \end{tabular}
    }
\label{tab:ablation-loss}
\end{table}
\begin{figure}[t]
  \centering
   \includegraphics[width=\linewidth]{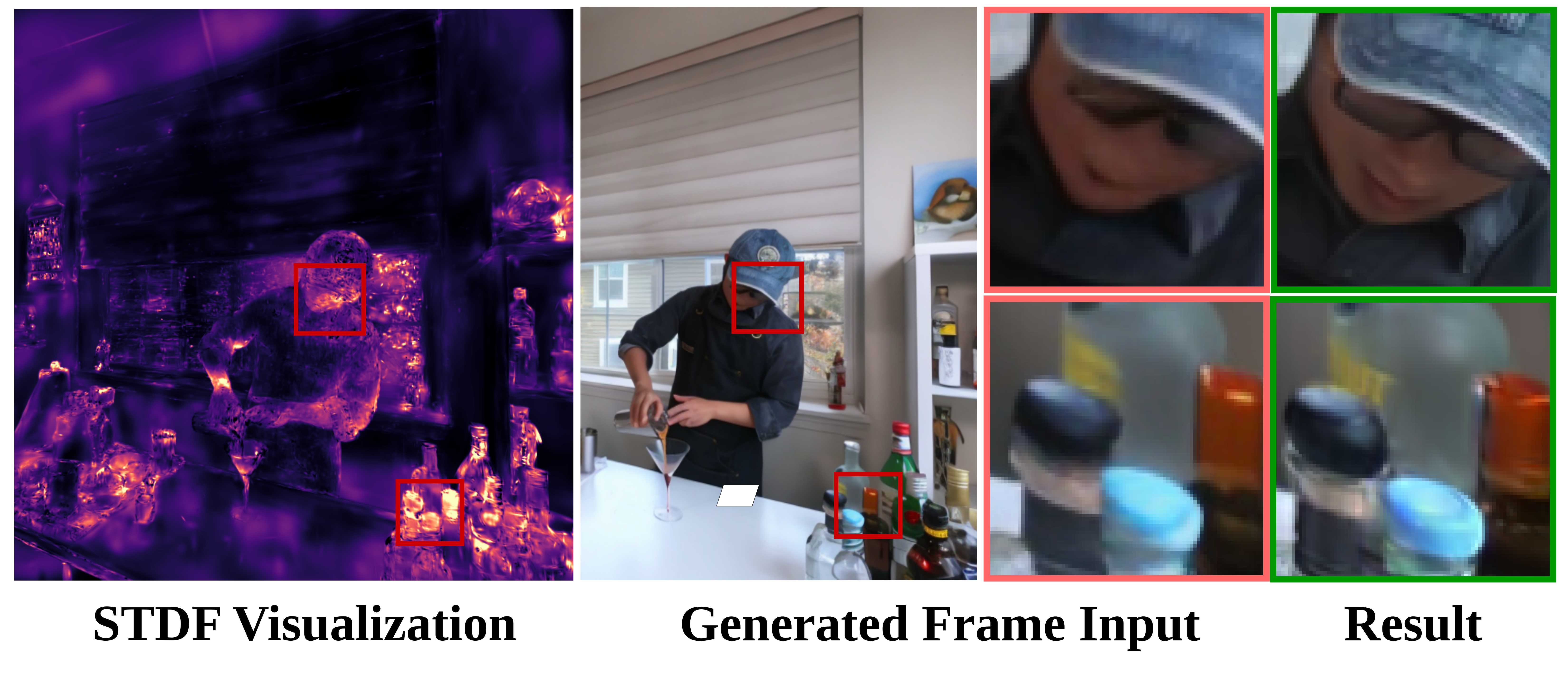}
   \vspace{-7mm}
   \caption{\textbf{Visualization of the STDF.} Spatio-Temporal Distortion Field output is rendered as a per-primitive attribute, with brighter regions indicating higher distortions (left). The corresponding areas in the input generated image (right) align with regions exhibiting noticeable deformation (red box).}   
   \label{fig:vis}
\end{figure}

\noindent{\textbf{Visualization of the Spatio-Temporal Distortion Field. }} We visualize the STDF output as a per-primitive attribute and render it with alpha-blending, as depicted in Fig.~\ref{fig:vis} (left). Heatmaps highlight regions with pronounced distortions in the generated observations, such as facial features and the wine bottle. This illustrates that, in the diffusion-generated observations along the trajectory, different contents undergo varying degrees of distortion, which relates to how the diffusion model perceives the physical world. After training, our method restores these regions in the GS rendering, demonstrating effective correction of spatio-temporal distortions (Fig.~\ref{fig:vis} (right)). For additional STDF visualization results, please refer to the supplementary material.

\begin{table}[t]
    \centering
    \captionof{table}{\textbf{Ablation study with alternative video diffusion models (VDMs) on Cook Spinach.}}
    \vspace{-2mm}
    \resizebox{0.9\linewidth}{!}{%
    \begin{tabular}{lccc}
        \toprule[1.5pt]
         Cook Spinach & PSNR$\uparrow$ & SSIM$\uparrow$ & LPIPS$\downarrow$ \\
        \midrule
            ViewCrafter   (w/o STDF) & 21.42 & 0.775 & 0.302 \\
            ViewCrafter   (w   STDF) & 23.93 & 0.832 & 0.232 \\
            ReCamMaster   (w/o STDF) & 21.97 & 0.756 & 0.315 \\
            ReCamMaster   (w   STDF) & 23.61 & 0.806 & 0.247 \\
        \bottomrule[1.5pt]
    \end{tabular}    
    }
    \label{tab:ablation-rcm}
\end{table}
\begin{figure}[t]
    \centering        
    \includegraphics[width=0.9\linewidth]{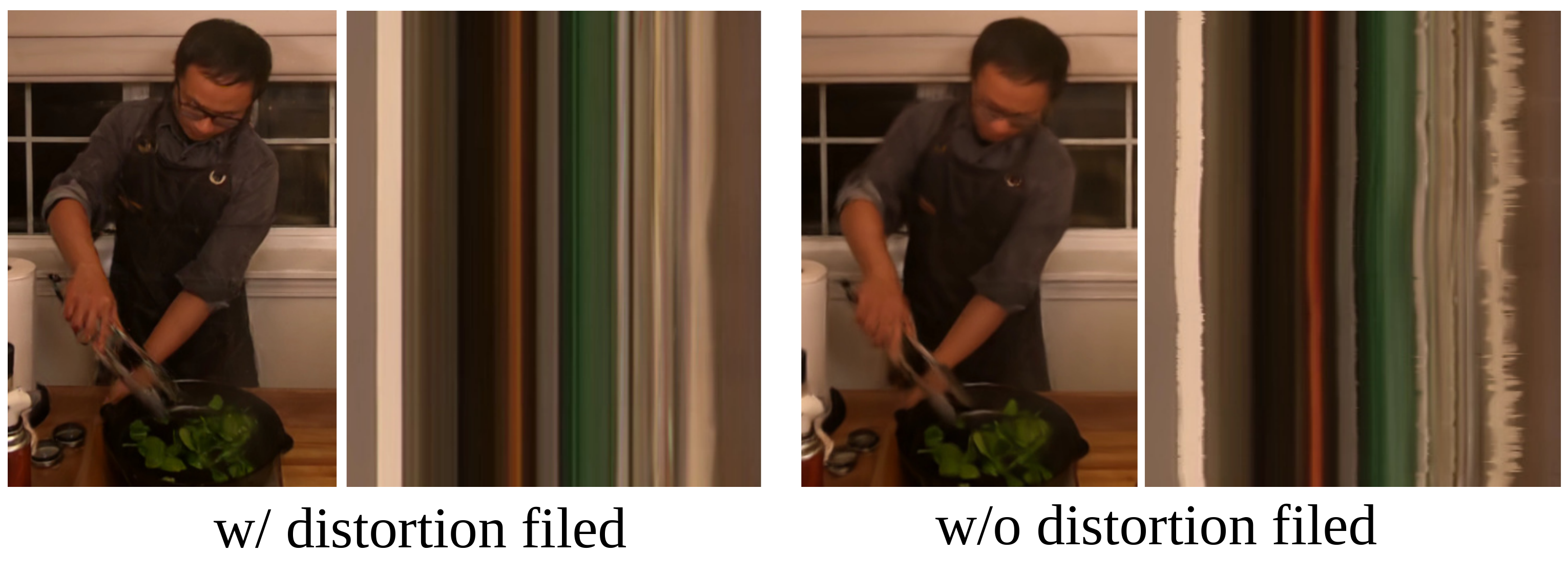}  
    \vspace{-2mm}
    \captionof{figure}{\textbf{Spatio-Temporal consistency on ReCamMaster.}}    
    \label{fig:ablation-rcm}   
\end{figure}

\noindent{\textbf{Generality Across Different Video Diffusion Models.}} As discussed earlier, inconsistencies in generated frames are inherent to existing VDMs. To further validate the applicability of our approach, we conduct experiment on another state-of-the-art camera-controlled VDM, \ie, ReCamMaster~\citep{bai2025recammaster}. 
As shown in Tab.~\ref{tab:ablation-rcm}, neither ViewCrafter nor ReCamMaster is able to directly reconstruct photorealistic 4D scenes, highlighting that inconsistencies in VDM-generated frames severely hinder the convergence of 4D content. In contrast, our method achieves significant improvements. When using ViewCrafter as the generative prior, our pipeline yields an increase of 2.51dB in PSNR, and when using ReCamMaster, a similar gain of 1.76db in PSNR is observed. These results demonstrate the capability of our approach to ensure spatio-temporal consistency even when conditioned on different generative models.

\section{Conclusion}

We propose a framework that leverages generative models for dynamic scene reconstruction from sparse cameras. We identify spatio-temporal inconsistencies in generative observations as the main barrier to achieving high-quality reconstruction. To address this, we introduce the Spatio-Temporal Distortion Field, explicitly modeling these inconsistencies across both space and time, and integrate it into a unified framework jointly optimizing pose, rendering, and smoothness for stable convergence. Both the main experiments and ablation studies confirm the advantage of our approach and highlight the critical role of unified spatio-temporal modeling. To the best of our knowledge, this is the first sparse-camera 4D reconstruction method thoroughly evaluated on standard multi-camera dynamic scene benchmarks, enabling widely accessible immersive 4D reconstruction.

\newpage

\noindent\textbf{Acknowledgment:} This work was partially supported by NSF of China (No. 62425209).

{
    \small
    \bibliographystyle{ieeenat_fullname}
    \bibliography{main}
}

\clearpage
\setcounter{page}{1}
\maketitlesupplementary

\section{Implementation Details}
Our framework is implemented in PyTorch, using the RealTime4DGS codebase as the foundation. To assess the versatility of our method, we did not apply dataset-specific tuning; instead, a unified training schedule was adopted across all datasets. 
For the optimization of 4D Gaussians, we strictly follow the official RealTime4DGS implementation. The Adam optimizer is employed throughout. The learning rate for each feature plane starts at $1.6\times10^{-3}$ and exponentially decays to $1.6\times10^{-4}$. The tiny MLP decoder is trained with an initial learning rate of $1.6\times10^{-4}$, which decays to $1.6\times10^{-5}$.
In the distortion field, we configure the base resolution of the spatial axes to 64. The basis resolution for temporal axis is set according to the total number of frames in the training sequence, while the pose axis is set to the number of generated views. For the multi-resolution feature structure, we employ a scale factor of 2 for each feature plane. We also find that expanding this factor from 2 to [2,4] will contribute to rendering quality at the expense of higher training cost.
To stabilize the optimization process, we introduce a warm-up stage during the first 1000 iterations. In this phase, only the 4D Gaussian primitives are optimized, while both the distortion field and camera pose optimization remain frozen. After this warm-up stage, joint optimization over the distortion field, camera poses, and Gaussian attributes is enabled. This staged training scheme prevents instability in the early iterations and ensures smoother convergence.

\subsection{Camera Selection}
\subsubsection{Training Cameras}
\begin{figure}[h]
    \centering
    \begin{minipage}{0.99\linewidth} 
        \centering             
        \includegraphics[width=\linewidth]{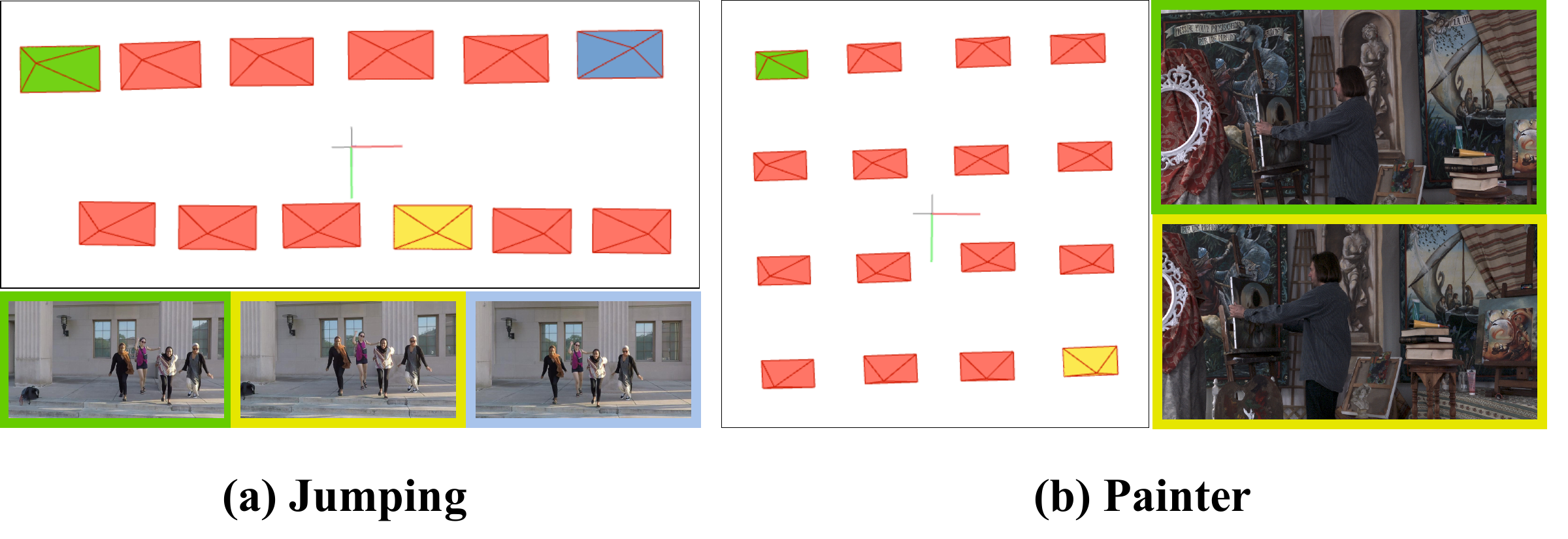}   
        \vspace{-4mm}
        \captionof{figure}{ 
          \textbf{Visualization of selected training cameras.} The non-red frustums denote the cameras used for training, corresponding to the images shown at the bottom and on the right respectively. The remaining red frustums indicate the cameras used for evaluation.}  
        \vspace{-2mm}
        \label{fig:supp-cam}
    \end{minipage}
\end{figure}

\begin{algorithm}[t]
\caption{Camera Subset Selection with Overlap-Constrained Greedy Expansion}
\label{alg:overlap_greedy}
\DontPrintSemicolon
\SetAlgoLined
\SetNoFillComment
\KwIn{\\
$P[i]$: Visible point sets for each camera $i$ \\
$\text{overlap}[i][j]$: overlap matrix between camera $i$ and camera $j$, defined as $\mathrm{overlap}[i][j]=\frac{|P_i \cap P_j|}{|P_i \cup P_j|}$\\
$o_{\min}$: overlap threshold \\
$\tau$: target coverage \\
$\mu$: overlap weight
}
\KwOut{Selected camera subset for training $C_s$}

\ForEach{camera $i$}{
    $G_{\text{nbr}}[i] \leftarrow \{ j \mid \text{overlap}[i][j] \ge o_{\min}, j \ne i \}$\;
}

$P_{\text{total}} \leftarrow \bigcup_i P[i]$ \tcp*{all visible points}
$U \leftarrow P_{\text{total}}$  \tcp*{uncovered points}

$c_0 \leftarrow \text{top-left camera}$\;
$C_s \gets \{c_0\}$\;
$U \gets U \setminus P[c_0]$\;
$c_{\text{cur}} \gets c_0$\;

\While{$1 - |U|/|P_{\text{total}}| < \tau$}{
    
    $\mathcal{N} \leftarrow \{ v \in G_{\text{nbr}}[c_{\text{cur}}] \mid v \notin C_s \}$\;
    
    \If{$\mathcal{N} = \emptyset$}{
        \textbf{break}\;
    }

    \ForEach{$v \in \mathcal{N}$}{
        $\text{gain}(v) \leftarrow |P[v] \cap U|$\;
        $\text{score}(v) \leftarrow \text{gain}(v) \cdot (1 + \mu \cdot \text{overlap}[c_{\text{cur}}][v])$\;
    }

    $v^\star \leftarrow \arg\max_{v \in \mathcal{N}} \text{score}(v)$\;

    \If{$|P[v^\star] \cap U| = 0$}{
        \textbf{break}\;
    }

    $C_s \gets C_s \cup \{v^\star\}$\;
    $U \gets U \setminus P[v^\star]$\;
    $c_{\text{cur}} \gets v^\star$\;
}

\Return{$C_s$}\;
\end{algorithm}

\begin{figure*}[t]
    \centering
    \begin{minipage}{0.98\linewidth} 
        \centering             
        \includegraphics[width=\linewidth]{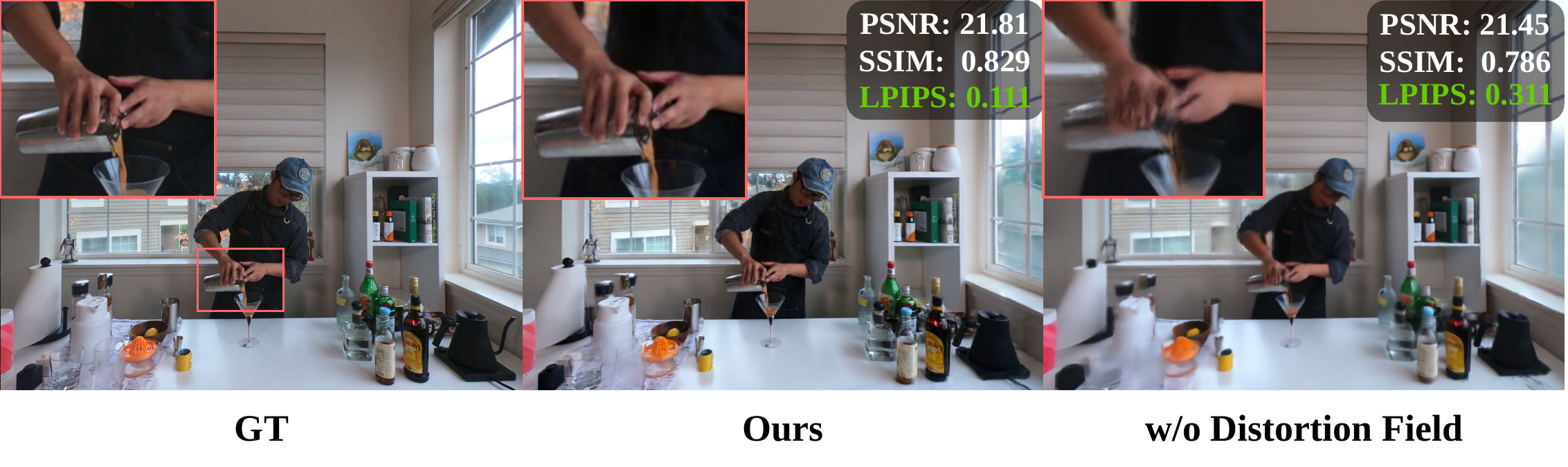}   
        \vspace{-8mm}
        \captionof{figure}{ 
         \textbf{Reliability of PSNR vs. LPIPS in Ablation Experiments.} We find that directly using generated-view reconstruction introduces severe oversmoothness that, paradoxically, favors PSNR computation, preventing PSNR from accurately reflecting changes in reconstruction quality. In contrast, LPIPS effectively captures the variations in reconstruction quality under these ablation settings.}  
        \vspace{-2mm}
        \label{fig:supp-metrics}
    \end{minipage}
\end{figure*}
We construct a minimal subset of cameras $C_s$ from the original set $C$ (usually containing 12-21 cameras) to serve as the training views. The subset $C_s$ is required to satisfy two conditions: 1) $C_s$ should sufficiently cover the scene content observed by $C$; 2) Since VDM will interpolate additional viewpoints within $C_s$, the per-frame overlap among cameras in $C_s$ is expected to be similar in order to ensure uniform scene coverage. Based on these criteria, we propose a greedy selection strategy that jointly considers visible point coverage and field-of-view overlap. The algorithm begins by choosing the camera located at the top-left position as the starting view. At each iteration, we construct a set of candidate cameras whose overlap with the current camera exceeds the minimum adjacency threshold $o_{min}$. For each candidate $v$, we compute a score that balances newly covered points and the geometric consistency with the current camera:
\begin{equation}
    \text{score}(v)=|P[v]\cap U|\cdot(1+\mu\cdot\text{overlap}[c_{cur}, v]),
\end{equation}
where $U$ denotes the set of currently uncovered points and $\mu=0.05$ is the overlap weight. The first term encourages maximal expansion of scene coverage, while the second ensures that consecutive selected cameras maintain sufficient overlap for reliable view interpolation using generative models. The candidate with the highest score is added to the selected subset, and the uncovered set is updated accordingly. The process iterates until the target coverage ratio is achieved or no valid candidates remain, yielding the final camera subset. We summarize the details in~\cref{alg:overlap_greedy}. Tab.~\ref{tab:supp-techni-ours}-\ref{tab:supp-nvidia-ours} provides camera index references for training and evaluation.~\cref{fig:supp-cam} illustrates the selected training cameras from two example scenes \textit{Jumping} and \textit{Painter}, with the training cameras highlighted in different colors.

\subsubsection{Evaluation Cameras}
To fully reflect the rendering quality of each method under the sparse-view setting, we do not adopt the original approach of evaluation using only one single selected camera. Instead, evaluation is performed on all remaining cameras.

\section{Metrics selection in Ablation Study}
In the ablation study, SSIM and LPIPS are adopted as the primary evaluation metrics instead of PSNR. This choice is motivated by two considerations. First, SSIM and LPIPS are more sensitive to structural details, with LPIPS in particular capturing perceptual differences in texture, sharpness, and local geometry, whereas PSNR primarily reflects pixel-wise color discrepancies and is less aligned with perceptual image quality. Second, as illustrated in~\cref{fig:supp-metrics}, the rendered images produced without distortion field exhibit substantial generative inconsistency, leading to overly smooth and blurred outputs. Such behavior tends to favor PSNR, resulting in only a marginal difference. In contrast, SSIM and LPIPS, especially LPIPS, capture the degradation of structural fidelity and perceptual quality, revealing the significant discrepancy.

\section{Additional Evaluation Results}

\subsection{Evaluation under Different Sparsity Levels.}
We compare our method with the baseline 4DGS approach, \ie 4DGaussian, under different sparsity levels, and the results are reported in~\cref{tab:supp-views}. Our method outperforms the baseline across all camera-view configurations, further demonstrating its practical robustness.
\begin{table}[h]
\centering
\caption{\textbf{Performance with different number of camera views.} We set 4DGaussian as the baseline model for dynamic scene reconstruction and evaluate the rendering quality with different number of training views.}
\resizebox{\linewidth}{!}{%
\begin{tabular}[t]{l|ccc|ccc|ccc}
\toprule
& \multicolumn{3}{c|}{3 views} & \multicolumn{3}{c|}{6 views} & \multicolumn{3}{c}{9 views} \\
& PSNR$\uparrow$ & SSIM$\uparrow$ & LPIPS$\downarrow$
& PSNR$\uparrow$ & SSIM$\uparrow$ & LPIPS$\downarrow$
& PSNR$\uparrow$ & SSIM$\uparrow$ & LPIPS$\downarrow$\\
\hline
4DGaussian  & 16.72 & 0.435 & 0.546 & 19.86 & 0.581 & 0.448 & 21.41 & 0.601 & 0.430  \\
Ours        & 21.56 & 0.656 & 0.264 & 23.58 & 0.687 & 0.243 & 24.72 & 0.751 & 0.205 \\

\bottomrule
\end{tabular}
}
\label{tab:supp-views}
\end{table}
\begin{figure*}[h]
    \centering
    \begin{minipage}{0.99\linewidth} 
        \centering
        \includegraphics[width=\linewidth]{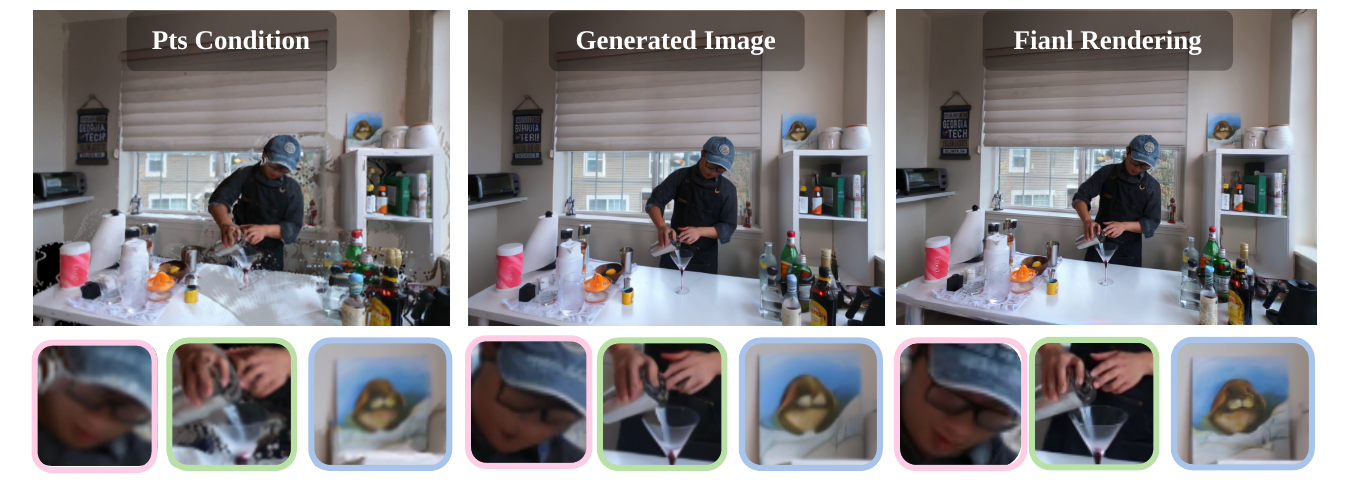}
        \vspace{-8mm}
        \captionof{figure}{ 
          Visualization of the point-cloud rendering condition, generation prior, and final results.}
        \vspace{-6mm}
        \label{fig:supp-gen}
    \end{minipage}
\end{figure*}
\subsection{Efficiency Comparison with RealTime4DGS}
Since our implementation is based on the RealTime4DGS codebase, we compare our method against it in terms of rendering fps to further evaluate efficiency. As shown in Tab~\ref{tab:supp-fps}, our approach maintains the similar rendering speed as RealTime4DGS.
Notably, the proposed distortion field is discarded after training, so inference complexity remains comparable to RealTime4DGS.
\begin{table}[h]
\centering
\caption{\textbf{Efficiency Comparison with RealTime4DGS.}}
\vspace{-2mm}
\resizebox{0.9\linewidth}{!}{%
\begin{tabular}[t]{lccc}
\toprule[1pt]
 & Coffee Martini & Playground & Average \\
\hline
RealTime4DGS & 11.48fps & 11.51fps & 11.50fps  \\
Ours & 11.82fps & 11.97fps & 11.90fps \\
\bottomrule[1pt]
\end{tabular}
}
\label{tab:supp-fps}
\vspace{-2mm}
\end{table}

\subsection{Per-scene Breakdown Results}
\label{supp:quantitative}
In Tab.~\ref{tab:supp-n3v}, Tab.~\ref{tab:supp-techni}, and Tab.~\ref{tab:supp-nvidia}, we provide a breakdown of the metrics for different scenes in the Neural 3D Video, Technicolor, and Nvidia Dynamic Scenes datasets, respectively. As shown in the quantitative results, our method demonstrates clear superiority over the other five approaches. This indicates that our method effectively leverages the generated priors while mitigating the inconsistency they may introduce. In addition, since our evaluation is conducted on all cameras excluding those used for training, we provide more detailed per-camera metrics, as reported in Tab.~\ref{tab:supp-techni-ours}-\ref{tab:supp-nvidia-ours}.

\subsection{Additional Qualitative Results}
\label{supp:qualitative}
\begin{figure*}[t]
    \centering
    \begin{minipage}{0.8\linewidth} 
        \centering             
        \includegraphics[width=\linewidth]{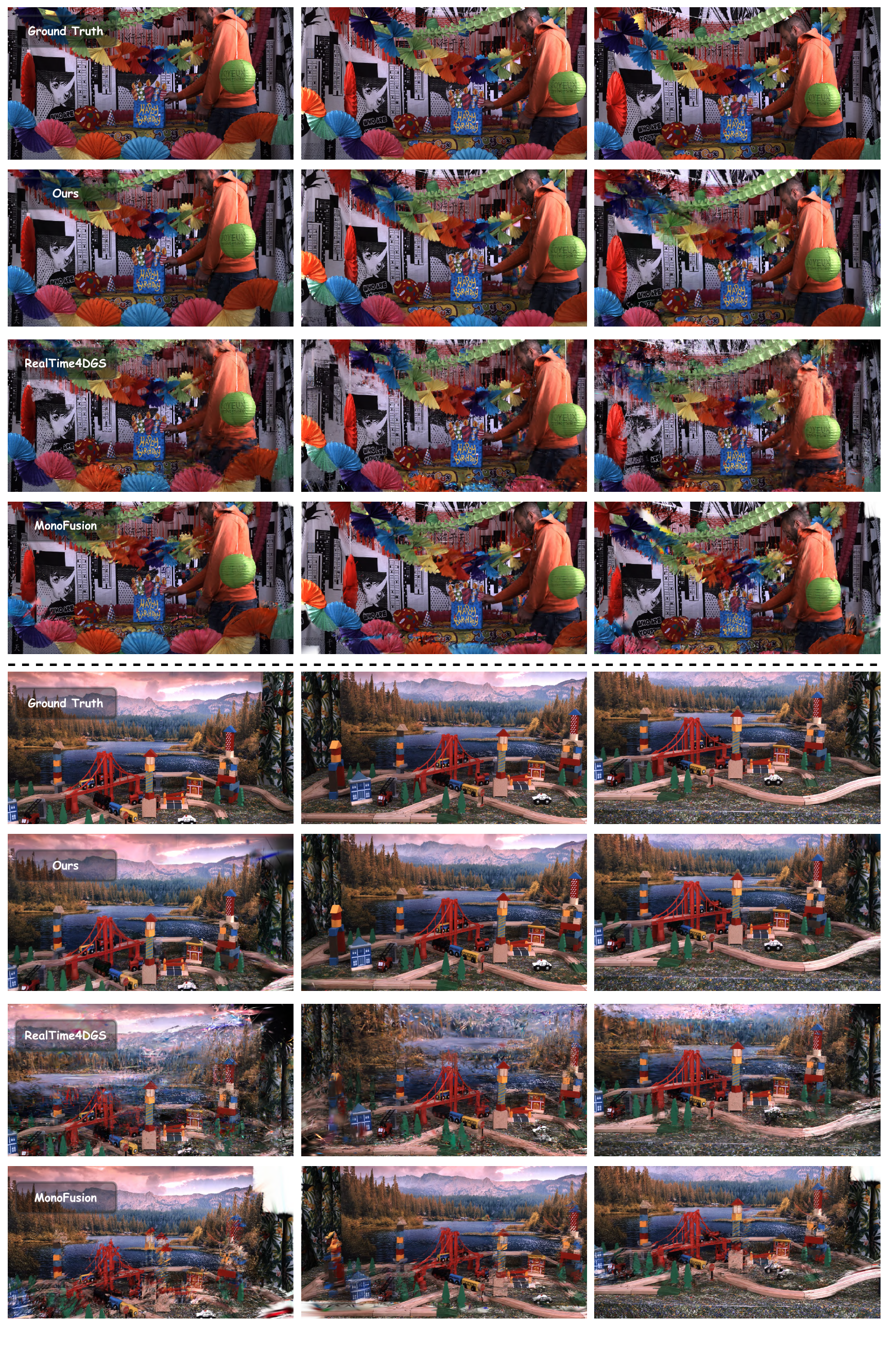}            
        \captionof{figure}{ 
          \textbf{Additional qualitative comparison results on \textit{Technicolor} Dataset.} We highlight the geometric completeness of background structures, bottom decorations, and the red bridge across different viewpoints, as well as the temporal consistency of dynamic regions such as human faces, the `Happy Birthday' text, and the toy train.}  
        \label{fig:supp-techni}
    \end{minipage}
\end{figure*}

\begin{figure*}[t]
    \centering
    \begin{minipage}{0.99\linewidth} 
        \centering             
        \includegraphics[width=\linewidth]{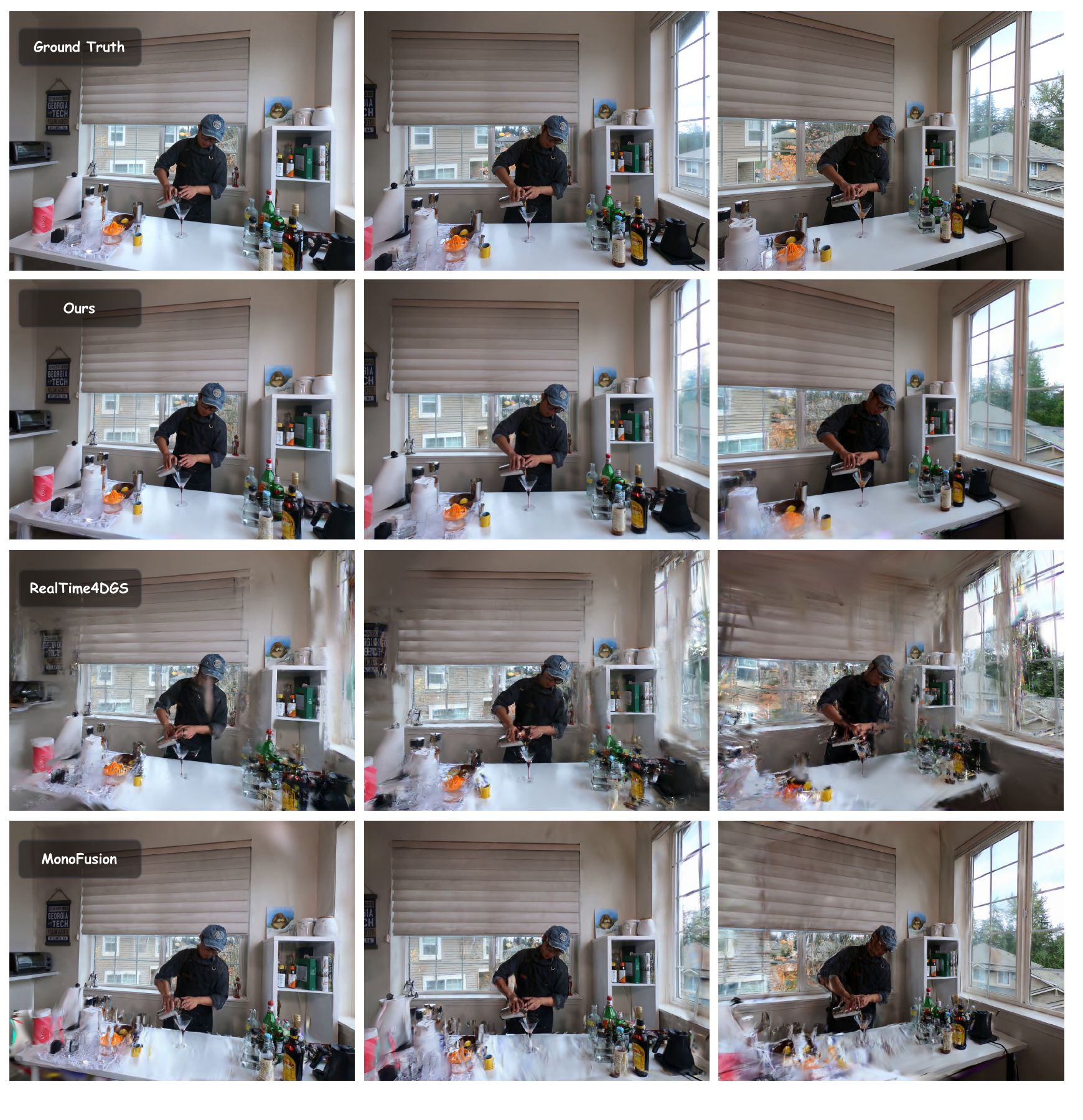}            
        \captionof{figure}{ 
          \textbf{Additional qualitative comparison results on \textit{Neural 3D Video} Dataset (part I).}}  
        \label{fig:supp-n3v-1}
    \end{minipage}
\end{figure*}

\begin{figure*}[t]
    \centering
    \begin{minipage}{0.99\linewidth} 
        \centering     
        \includegraphics[width=\linewidth]{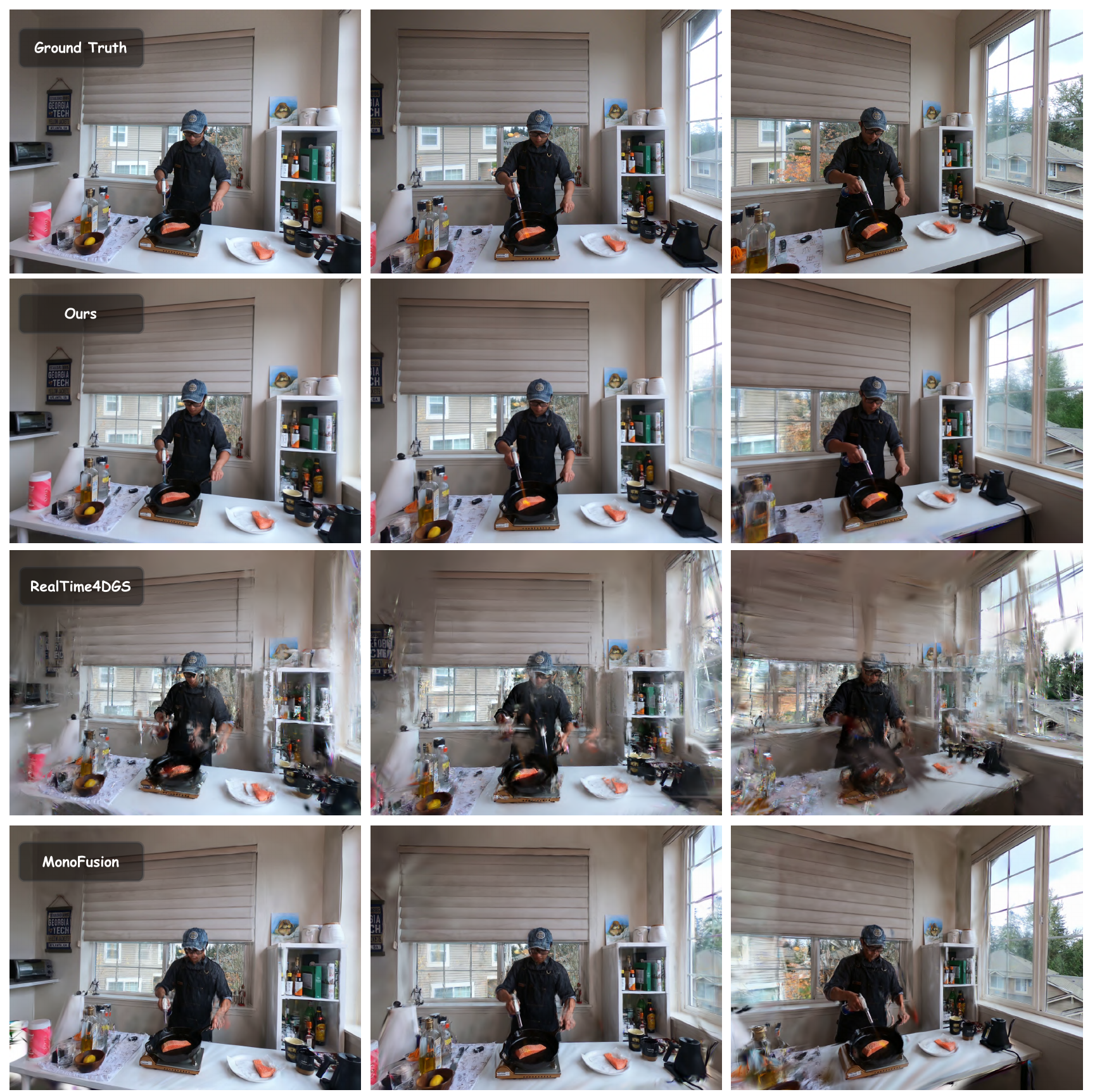}            
        \captionof{figure}{ 
          \textbf{Additional qualitative comparison results on \textit{Neural 3D Video} Dataset (part II).}}  
        \label{fig:supp-n3v-2}
    \end{minipage}
\end{figure*}

\begin{figure*}[t]
    \centering
    \begin{minipage}{0.99\linewidth} 
        \centering     
        \includegraphics[width=\linewidth]{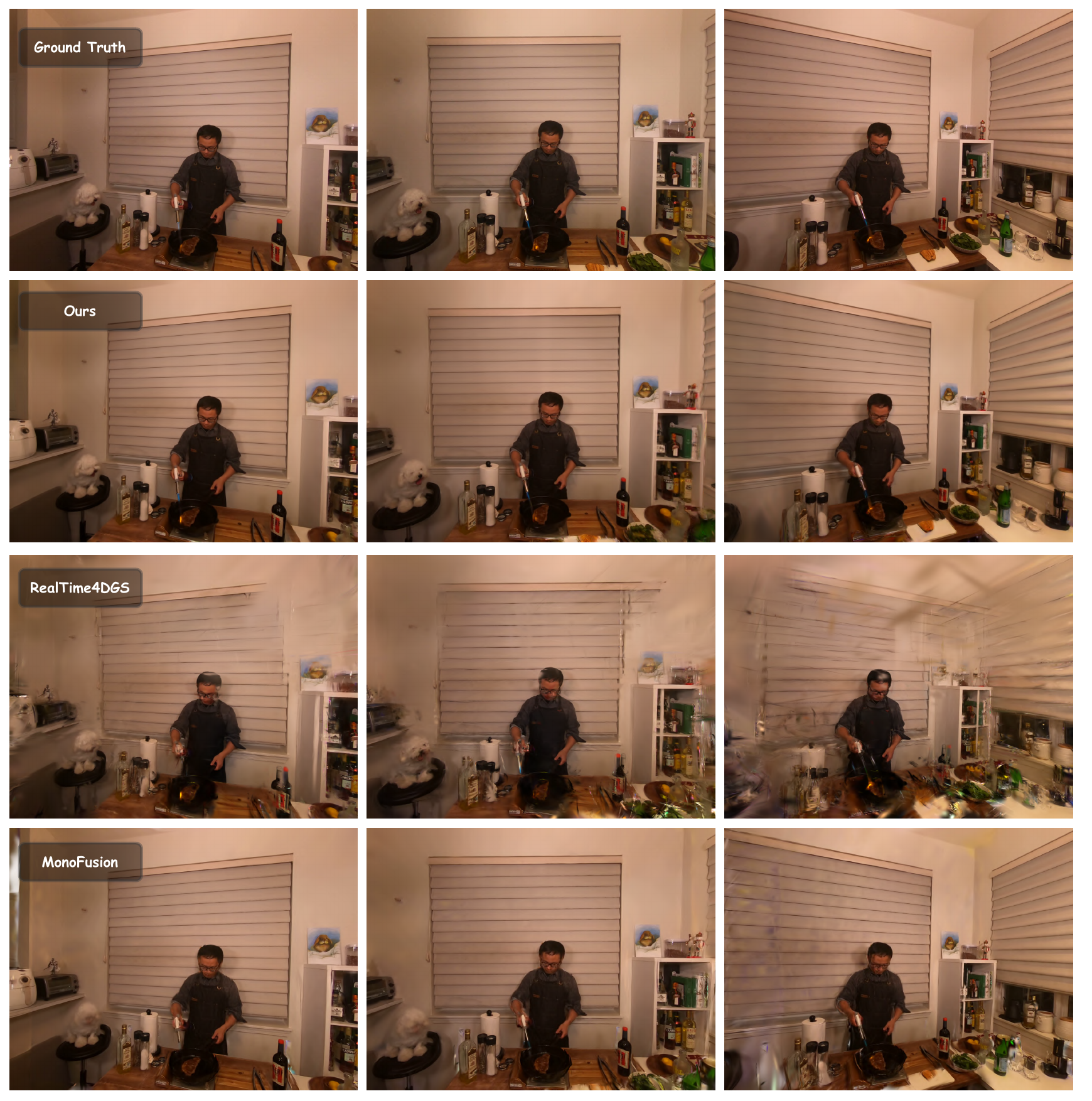}            
        \captionof{figure}{ 
          \textbf{Additional qualitative comparison results on \textit{Neural 3D Video} Dataset (part III).}}  
        \label{fig:supp-n3v-3}
    \end{minipage}
\end{figure*}

\begin{figure*}[t]
    \centering
    \begin{minipage}{0.99\linewidth} 
        \centering     
        \includegraphics[width=\linewidth]{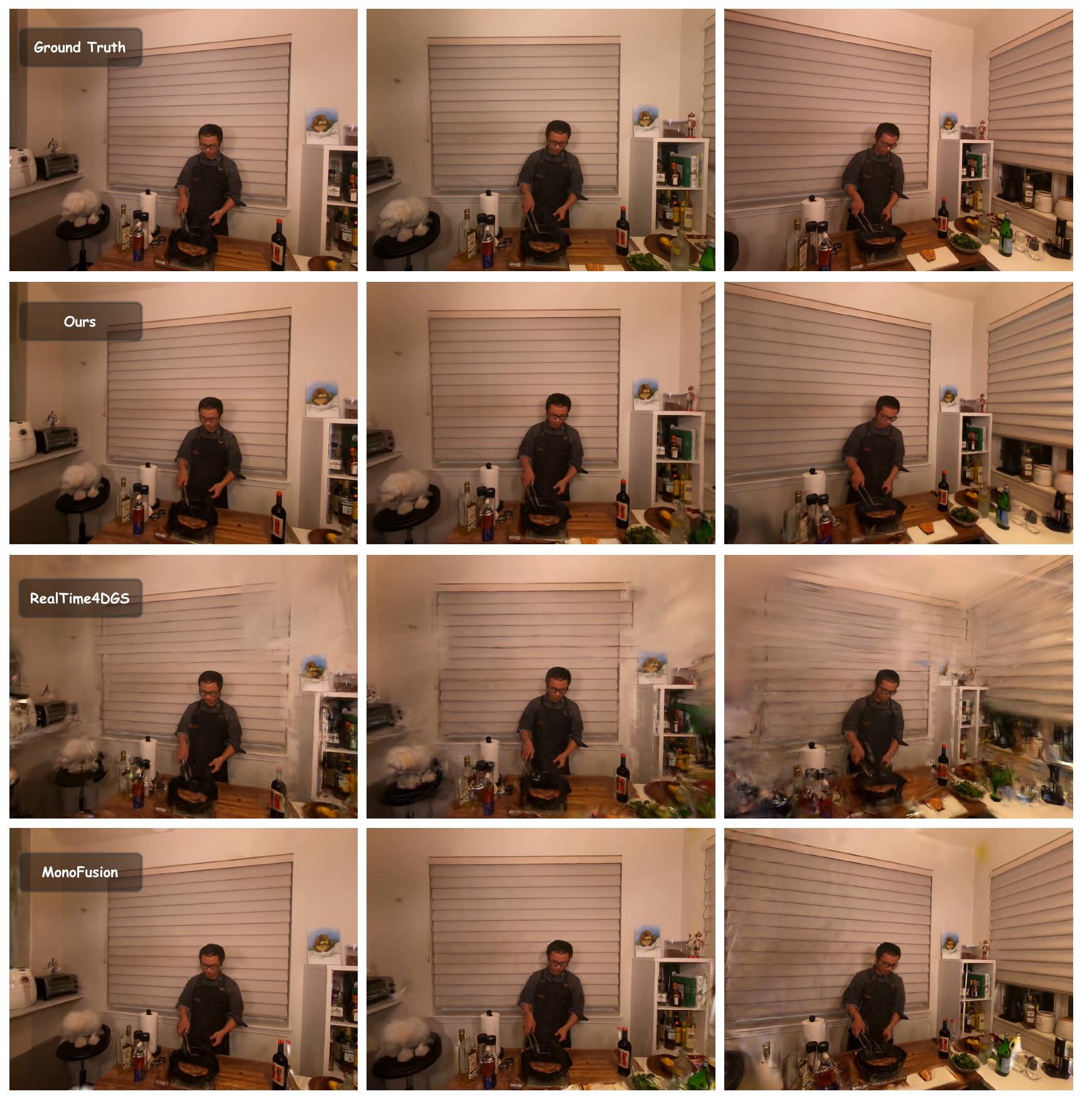}  
        \captionof{figure}{ 
          \textbf{Additional qualitative comparison results on \textit{Neural 3D Video} Dataset (part IV).}}  
        \label{fig:supp-n3v-4}
    \end{minipage}
\end{figure*}

\begin{figure*}[t]
    \centering
    \begin{minipage}{0.8\linewidth} 
        \centering      
        \includegraphics[width=\linewidth]{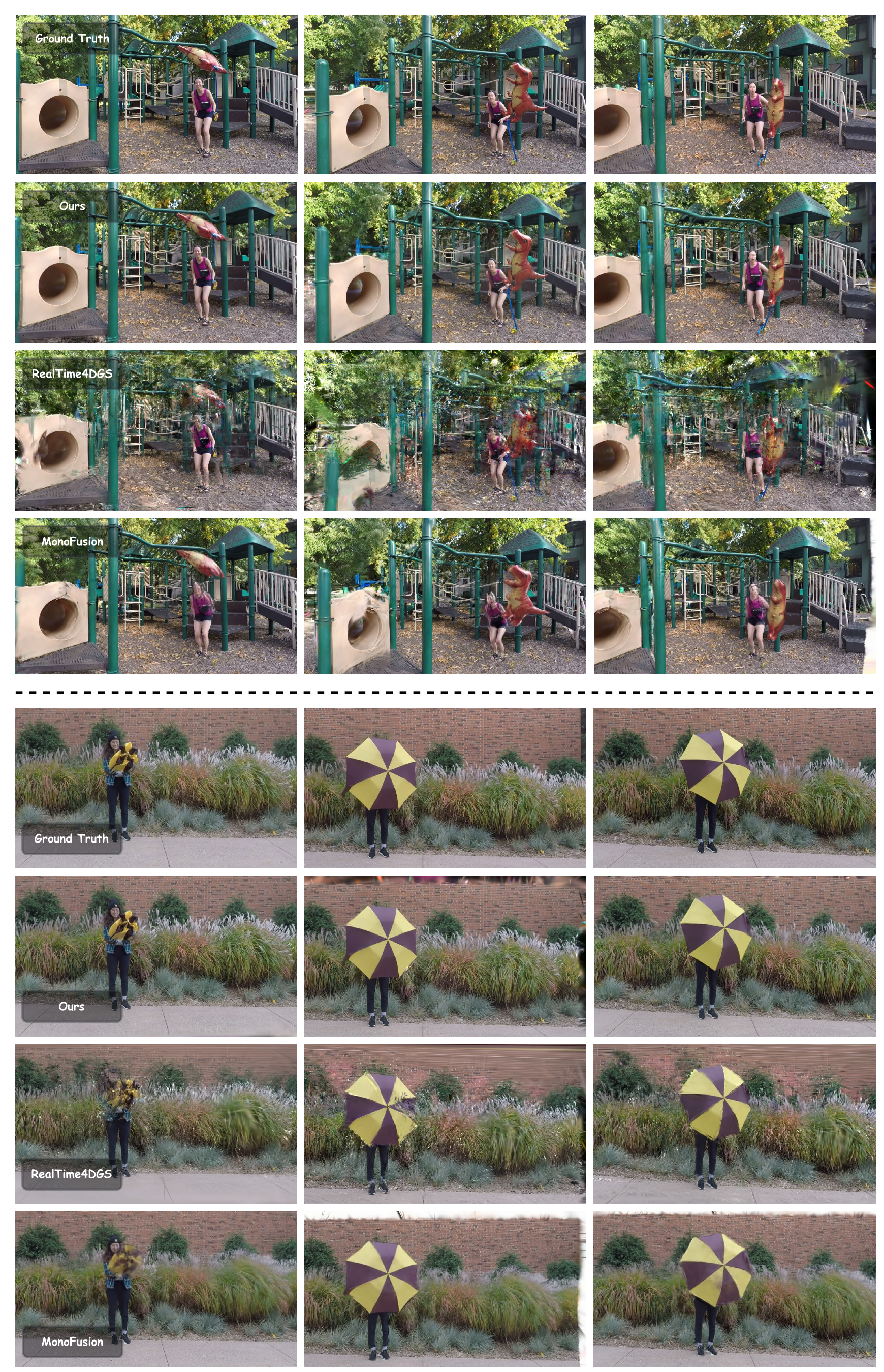} 
        \captionof{figure}{ 
          \textbf{Additional qualitative comparison results on \textit{Nvidia Dynamic Scenes} Dataset (part I).} In the top panel, we emphasize the geometric completeness of the left archway and the right escalator across different viewpoints, as well as the temporal consistency of dynamic regions such as the human body, balloon, and the blue tether. In the bottom panel, we highlight the fine details of background grass, shoes, and umbrella textures, together with accurate motion fitting under fast movement in the leftmost column.}  
        \label{fig:supp-nvidia-2}
    \end{minipage}
\end{figure*}

\begin{figure*}[t]
    \centering
    \begin{minipage}{0.8\linewidth} 
        \centering      
        \includegraphics[width=\linewidth]{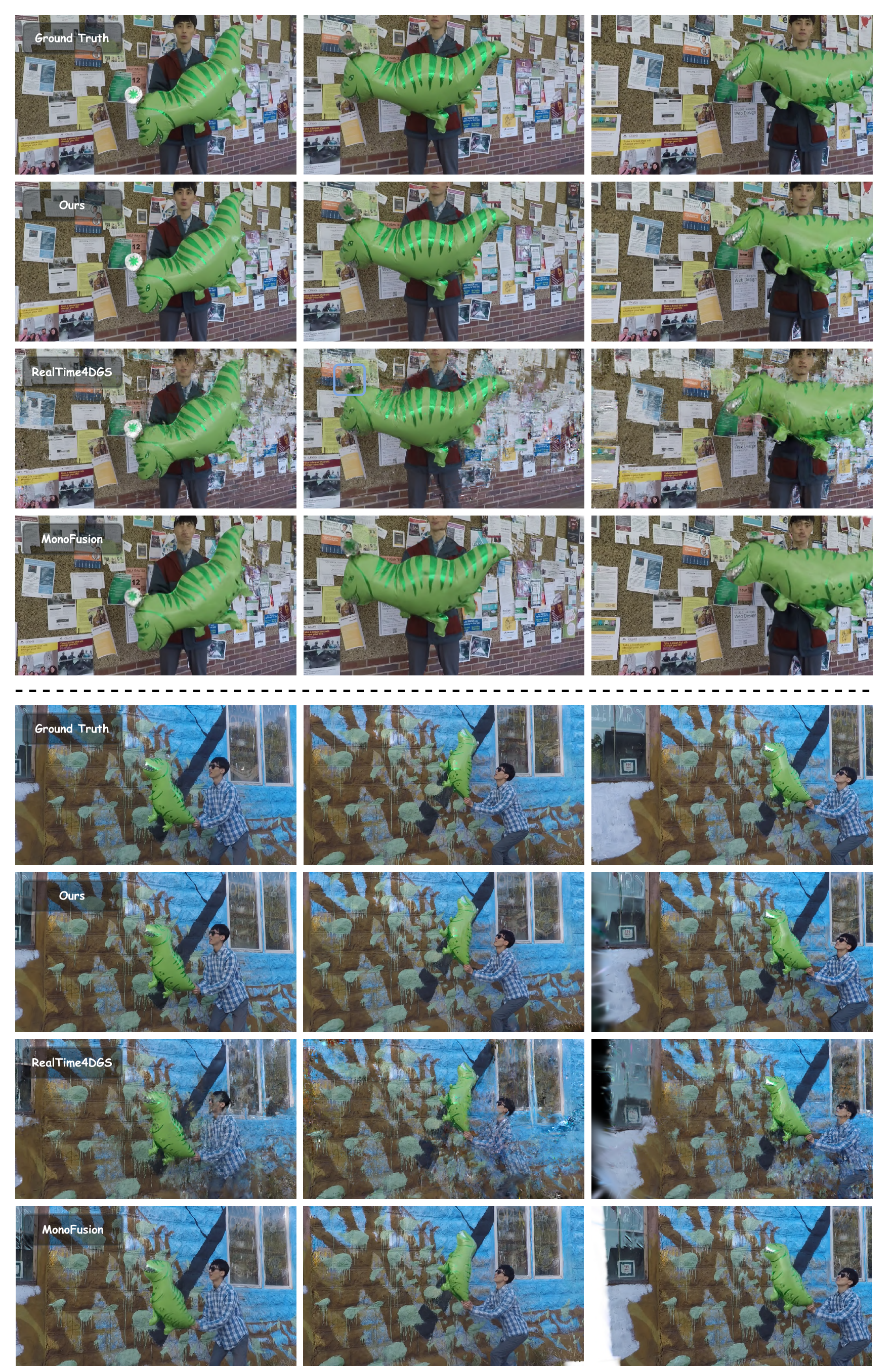}                   
        \captionof{figure}{ 
          \textbf{Additional qualitative comparison results on \textit{Nvidia Dynamic Scenes} Dataset (part II).}}  
          \label{fig:supp-nvidia-1}
    \end{minipage}
\end{figure*}

In Fig.~\ref{fig:supp-techni}, Fig.~\ref{fig:supp-n3v-1}-\ref{fig:supp-n3v-4}, and Fig.~\ref{fig:supp-nvidia-1},~\ref{fig:supp-nvidia-2}, we showcase additional rendering comparisons with those methods aforementioned at different timesteps. Our method consistently recovers high-quality and
coherent renderings over time, demonstrating its effectiveness in dynamic scenes with sparse observation. Please zoom in for more details.

\begin{figure*}[t]
    \centering
    \begin{minipage}{0.99\linewidth} 
        \centering             
        \includegraphics[width=\linewidth]{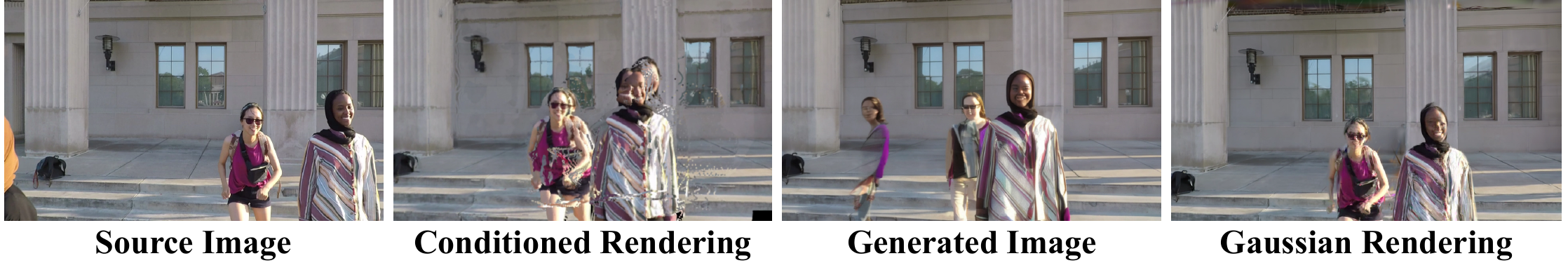}   
        \vspace{-6mm}
        \captionof{figure}{ 
          \textbf{Visualization of one failure case.} From left to right are the source image and point-cloud–rendering condition provided to ViewCrafter, the corresponding generated image, and a Gaussian rendering near that generated image. Since ViewCrafter is primarily trained on scene-centric data with few human subjects, this example suffers from an out-of-domain issue. In addition, the quality of the point-cloud–rendering condition is relatively low. These factors jointly lead to low-quality generated images, which in turn degrade the Gaussian reconstruction quality on the human body.}  
        \vspace{-2mm}
        \label{fig:supp-failure}
    \end{minipage}
\end{figure*}

\section{More Details for Generated Images}
In our main experiments, we adopt ViewCrafter to provide additional observations, from which 20–25 generated views are uniformly sampled for training. ViewCrafter conditions its diffusion model on point-cloud renderings obtained from Dust3R, given two input images along with the target trajectory. However, we find that the quality of generated images is highly dependent on the quality of the point-cloud renderings. To improve the generation quality, we replace Dust3r with a stronger 3D foundation model, VGGT, to produce more reliable conditioning signals. Despite this improvement, the generated sequences still exhibit spatio-temporal inconsistencies that hinder their direct use for 4D reconstruction, as illustrated in Fig~\ref{fig:supp-gen}. Such inconsistencies can lead to severe over-blurring in the reconstructed results, as shown in Fig.~\ref{fig:supp-metrics} (right).

As for the ablation study with ReCamMaster, unlike ViewCrafter, which leverages point-cloud renderings as trajectory priors, ReCamMaster directly conditions on camera parameters to generate a video sequence of 81 frames. Without explicit geometric constraints, the generated images from vanilla ReCamMaster suffer from severe hallucinations, making them unsuitable for reconstruction. To address this, we finetune ReCamMaster by incorporating point-cloud renderings as an additional conditioning signal, analogous to ViewCrafter. 

\begin{figure*}[h]
    \centering
    \begin{minipage}{0.99\linewidth} 
        \centering             
        \includegraphics[width=\linewidth]{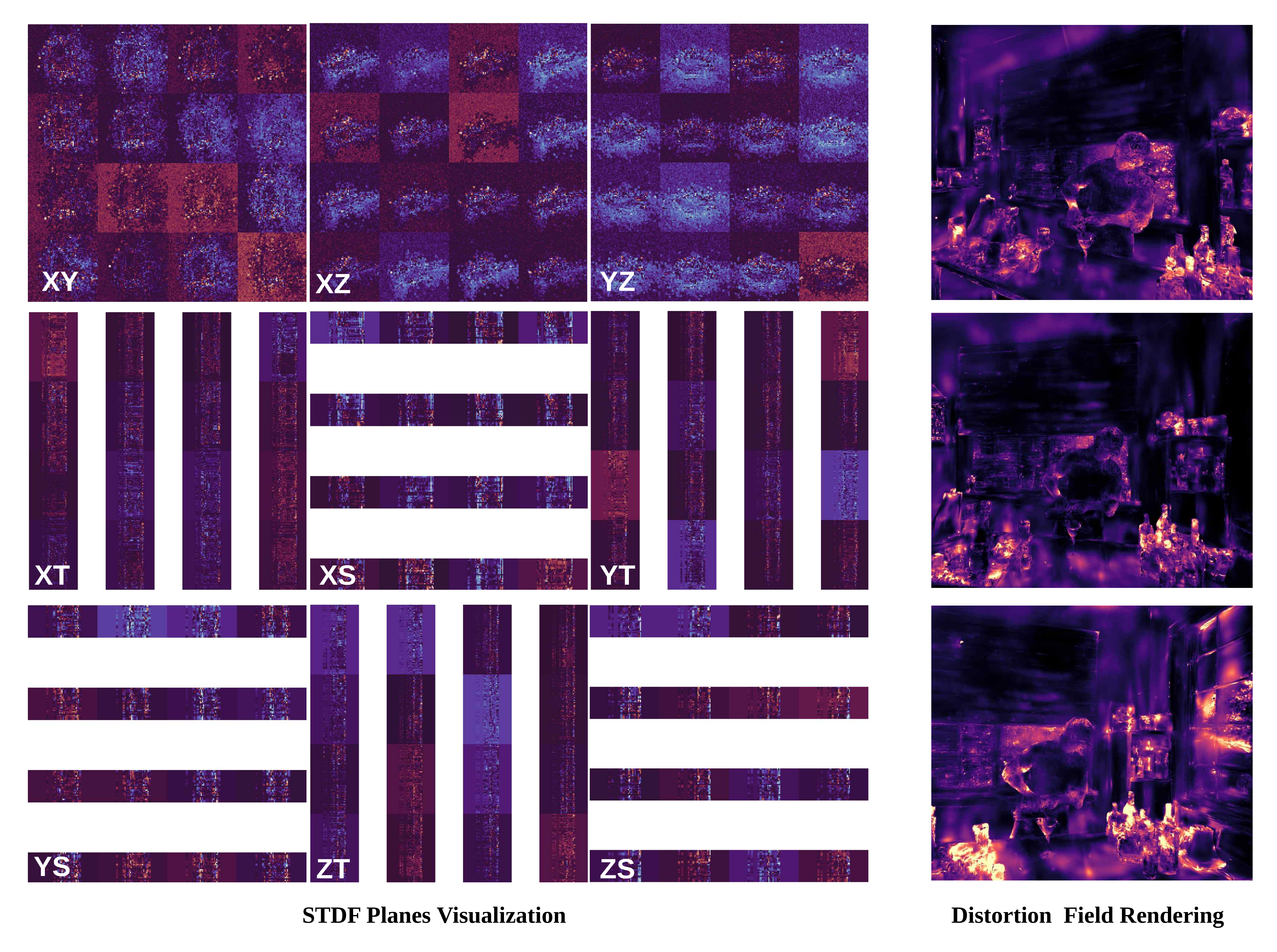}   
        \vspace{-8mm}
        \captionof{figure}{ 
          Visualization of Spatio-Temporal Distortion Field}  
        \vspace{-4mm}
        \label{fig:supp-field}
    \end{minipage}
\end{figure*}
\subsection{Failure Case}
Although our method substantially mitigates the inter-frame inconsistencies caused by distortions in generated images, the final performance still partially depends on the quality of the generated images themselves. As illustrated in~\cref{fig:supp-failure}, when the generated images are of poor quality—typically due to out-of-domain inputs or generative hallucinations—such as the severely deformed human body and the additional ghosting artifacts shown in the figure, addressing only the inter-image inconsistency is insufficient to achieve satisfactory reconstruction results. Without considering training cost, incorporating the generative model into the reconstruction pipeline for joint or iterative optimization is a promising direction for future exploration.

\section{More visualization}
To gain an intuitive understanding of the STDF training results, we visualize the full feature maps (16 channels) of each plane in the CoffeMartini scene in~\cref{fig:supp-field}. The activated regions vary across different dimensions, indicating the intertwined nature of spatial and temporal deformations.

To further interpret the outputs, we render the deformation predictions of STDF as an additional attribute field. This allows us to clearly observe the regions in the scene where distortions occur. Through this visualization approach, we can intuitively perceive the spatio-temporal distortions present in the 4D scene.
\newpage
\clearpage
\begin{table}[H]
\centering
\caption{PSNR, SSIM, and LPIPS of our framework on Technicolor, with cam00, cam09 and cam15 served as training views.}
\resizebox{0.98\linewidth}{!}{%
\begin{tabular}[t]{lcccccc}
\toprule[1pt]
 & \multicolumn{5}{c}{\textbf{Technicolor - PSNR}} \\
\midrule
 & Fabien & Theater & Train & Birthday & Painter & Average \\
\hline\addlinespace[0.5ex]
HyperReel       &13.62&14.24&15.44&12.28&15.13&14.14 \\
4DGaussians     &18.31&17.53&16.72&11.97&16.48&16.20 \\
4DRotor         &15.21&16.00&14.01&12.42&16.62&14.85 \\
RealTime4DGS    &13.00&16.27&16.59&18.45&18.76&16.53 \\
MonoFusion      &17.86&19.48&15.12&16.47&20.92&17.97 \\
\hline\addlinespace[0.5ex]
Ours            &25.23&22.44&21.56&20.11&26.39&23.15 \\
\bottomrule[1pt]
\vspace{1mm} 
\end{tabular}
}
\resizebox{0.98\linewidth}{!}{%
\begin{tabular}[t]{lcccccc}
\toprule[1pt]
 & \multicolumn{5}{c}{\textbf{Technicolor - SSIM}} \\
\midrule
 & Fabien & Theater & Train & Birthday & Painter & Average \\
\hline\addlinespace[0.5ex]
HyperReel       &0.653&0.434&0.351&0.341&0.484&0.453 \\
4DGaussians     &0.704&0.552&0.435&0.335&0.499&0.505\\
4DRotor         &0.625&0.479&0.270&0.307&0.450&0.426 \\
RealTime4DGS    &0.520&0.451&0.438&0.586&0.556&0.510 \\
MonoFusion      &0.737&0.590&0.392&0.524&0.648&0.578 \\
\hline\addlinespace[0.5ex]
Ours            &0.814&0.686&0.656&0.683&0.802&0.728 \\
\bottomrule[1pt]
\vspace{1mm}
\end{tabular}
}
\resizebox{0.98\linewidth}{!}{%
\begin{tabular}[t]{lcccccc}
\toprule[1pt]
 & \multicolumn{5}{c}{\textbf{Technicolor - LPIPS}} \\
\midrule
 & Fabien & Theater & Train & Birthday & Painter & Average \\
\hline\addlinespace[0.5ex]
HyperReel       &0.514&0.666&0.595&0.664&0.642&0.616 \\
4DGaussians     &0.492&0.539&0.546&0.620&0.564&0.552 \\
4DRotor         &0.574&0.571&0.589&0.612&0.560&0.581 \\
RealTime4DGS    &0.614&0.623&0.505&0.437&0.529&0.542 \\
MonoFusion      &0.415&0.380&0.356&0.339&0.271&0.352 \\
\hline\addlinespace[0.5ex]
Ours            &0.340&0.379&0.264&0.274&0.236&0.299 \\
\bottomrule[1pt]
\end{tabular}
}
\label{tab:supp-techni}
\end{table}

\FloatBarrier
\begin{table}[H]
\centering
\caption{PSNR, SSIM, and LPIPS of our framework on Neural 3D Video, with cam01, cam05 and cam10 served as training views.}
\resizebox{0.98\linewidth}{!}{%
\begin{tabular}[t]{lccccccc}
\toprule[2pt]
 & \multicolumn{6}{c}{\textbf{Neural 3D Video - PSNR}} \\
\midrule
 & coffee martini & cook spinach & cut roasted beef & flame salmon & flame steak & sear steak & Average \\
\hline\addlinespace[0.5ex]
HyperReel       &14.07&16.44&17.04&10.64&17.57&17.28&15.63 \\
4DGaussians     &14.49&17.94&18.34&14.83&19.53&18.57&17.40 \\
4DRotor         &15.86&19.65&18.51&16.10&19.69&19.05&18.20 \\
RealTime4DGS    &15.07&17.95&17.04&14.65&19.57&18.86&17.31 \\
MonoFusion      &15.59&19.71&19.15&15.42&20.02&19.91&18.43 \\
\hline\addlinespace[0.5ex]
Ours            &18.40&23.93&22.43&18.92&23.30&23.56&21.91 \\
\bottomrule[2pt]
\vspace{1mm}
\end{tabular}
}

\resizebox{0.98\linewidth}{!}{%
\begin{tabular}[t]{lccccccc}
\toprule[2pt]
 & \multicolumn{6}{c}{\textbf{Neural 3D Video - SSIM}} \\
\midrule
 & coffee martini & cook spinach & cut roasted beef & flame salmon & flame steak & sear steak & Average \\
\hline\addlinespace[0.5ex]
HyperReel       &0.520&0.586&0.635&0.452&0.644&0.633&0.582 \\
4DGaussian      &0.593&0.664&0.696&0.611&0.736&0.716&0.673 \\
4DRotor         &0.649&0.741&0.715&0.651&0.753&0.744&0.708 \\
RealTime4DGS    &0.590&0.670&0.640&0.569&0.708&0.700&0.649 \\
MonoFusion      &0.643&0.780&0.777&0.630&0.788&0.784&0.738 \\
\hline\addlinespace[0.5ex]
Ours            &0.729&0.832&0.792&0.726&0.820&0.818&0.789 \\
\bottomrule[2pt]
\vspace{1mm}
\end{tabular}
}

\resizebox{0.98\linewidth}{!}{%
\begin{tabular}[t]{lccccccc}
\toprule[2pt]
 & \multicolumn{6}{c}{\textbf{Neural 3D Video - LPIPS}} \\
\midrule
 & coffee martini & cook spinach & cut roasted beef & flame salmon & flame steak & sear steak & Average \\
\hline\addlinespace[0.5ex]
HyperReel       &0.540&0.499&0.447&0.604&0.466&0.459&0.500 \\
4DGaussians     &0.401&0.320&0.325&0.362&0.258&0.270&0.320 \\
4DRotor         &0.406&0.334&0.359&0.371&0.334&0.336&0.357 \\
RealTime4DGS    &0.471&0.428&0.459&0.485&0.405&0.412&0.442 \\
MonoFusion      &0.367&0.228&0.236&0.354&0.229&0.231&0.270 \\
\hline\addlinespace[0.5ex]
Ours            &0.298&0.232&0.259&0.291&0.237&0.238&0.258 \\
\bottomrule[2pt]
\vspace{1mm}
\end{tabular}
}
\label{tab:supp-n3v}
\end{table}

\FloatBarrier
\begin{table}[H]
\centering
\caption{PSNR, SSIM, and LPIPS of our framework on Nvidia Dynamic Scenes, with cam01, cam06 and cam10 as training views.}
\resizebox{0.98\linewidth}{!}{%
\begin{tabular}[t]{lccccccc}
\toprule[2pt]
 & \multicolumn{6}{c}{\textbf{Nvidia Dynamic Scenes - PSNR}} \\
\midrule
 & jumping & balloon1 & balloon3 & playground & skating & umbrella & Average \\
\hline\addlinespace[0.5ex]
HyperReel       &24.45&15.46&22.58&15.41&21.05&24.74&19.88 \\
4DGaussians     &17.38&16.96&14.56&12.75&19.70&19.52&16.81 \\
4DRotor         &21.90&14.55&20.33&10.95&26.28&22.29&19.38 \\
RealTime4DGS    &17.85&18.36&19.07&13.96&18.52&19.71&17.91 \\
MonoFusion      &20.72&20.31&19.81&16.86&23.55&20.07&20.22 \\
\hline\addlinespace[0.5ex]
Ours            &24.36&25.31&25.33&19.64&28.73&25.42&24.81 \\
\bottomrule[2pt]
\end{tabular}
}
\vspace{2mm} \\
\resizebox{0.98\linewidth}{!}{%
\begin{tabular}[t]{lccccccc}
\toprule[2pt]
 & \multicolumn{6}{c}{\textbf{Nvidia Dynamic Scenes - SSIM}} \\
\midrule
 & jumping & balloon1 & balloon3 & playground & skating & umbrella & Average \\
\hline\addlinespace[0.5ex]
HyperReel       &0.782&0.350&0.632&0.281&0.669&0.629&0.528 \\
4DGaussians     &0.525&0.372&0.264&0.166&0.607&0.299&0.372 \\
4DRotor         &0.731&0.311&0.497&0.137&0.814&0.556&0.508 \\
RealTime4DGS    &0.585&0.533&0.509&0.267&0.649&0.333&0.479 \\
MonoFusion      &0.687&0.563&0.542&0.499&0.819&0.428&0.590 \\
\hline\addlinespace[0.5ex]
Ours            &0.793&0.810&0.791&0.672&0.906&0.794&0.794 \\
\bottomrule[2pt]
\end{tabular}
}
\vspace{2mm} \\
\resizebox{0.98\linewidth}{!}{%
\begin{tabular}[t]{lccccccc}
\toprule[2pt]
 & \multicolumn{6}{c}{\textbf{Nvidia Dynamic Scenes - LPIPS}} \\
\midrule
 & jumping & balloon1 & balloon3 & playground & skating & umbrella & Average \\
\hline\addlinespace[0.5ex]
HyperReel       &0.256&0.537&0.266&0.491&0.450&0.238&0.396 \\
4DGaussians     &0.454&0.473&0.609&0.608&0.449&0.498&0.516 \\
4DRotor         &0.245&0.601&0.392&0.670&0.166&0.260&0.389 \\
RealTime4DGS    &0.450&0.371&0.380&0.499&0.422&0.432&0.426 \\
MonoFusion      &0.241&0.151&0.174&0.206&0.134&0.248&0.192 \\
\hline\addlinespace[0.5ex]
Ours            &0.170&0.127&0.142&0.190&0.121&0.148&0.150 \\
\bottomrule[2pt]
\end{tabular}
}
\label{tab:supp-nvidia}
\end{table}
\begin{table}[H]
\centering
\caption{\textbf{PSNR, SSIM, and LPIPS of our framework on Technicolor.} Three cameras (cam00, cam09 and cam15) are served as training views for scene \textit{Fabien}, \textit{Theater} and \textit{Train}, while only two cameras (cam00 and cam15) are training views for scene \textit{Birthday} and \textit{Painter}.}
\tiny
\resizebox{0.98\linewidth}{!}{%
\begin{tabular}[t]{lcccccc}
\toprule[1pt]
 & \multicolumn{5}{c}{\textbf{Technicolor - PSNR}} \\
\midrule
 & Fabien & Theater & Train & Birthday & Painter & Average \\
\hline
cam01 & 25.581 & 21.231 & 23.456 & 19.562 & 26.322 & 23.23 \\
cam02 & 21.886 & 19.203 & 21.202 & 18.227 & 25.736 & 21.251 \\
cam03 & 18.844 & 18.5 & 19.468 & 17.692 & 27.939 & 20.488 \\
cam04 & 28.676 & 23.089 & 23.59 & 20.991 & 27.423 & 24.754 \\
cam05 & 28.162 & 23.422 & 23.833 & 20.837 & 27.249 & 24.701 \\
cam06 & 23.838 & 20.949 & 21.0 & 19.824 & 25.58 & 22.238 \\
cam07 & 20.735 & 19.916 & 19.228 & 19.306 & 25.717 & 20.98 \\
cam08 & 28.175 & 23.997 & 23.649 & 20.245 & 25.817 & 24.377 \\
cam10 & 25.706 & 23.722 & 20.932 & 20.962 & 26.613 & 23.587 \\
cam11 & 24.939 & 23.549 & 21.013 & 21.682 & 27.519 & 23.741 \\
cam12 & 25.122 & 23.568 & 20.321 & 19.482 & 23.795 & 22.457 \\
cam13 & 27.505 & 25.954 & 21.626 & 20.643 & 26.436 & 24.433 \\
cam14 & 28.818 & 24.666 & 20.962 & 21.986 & 26.913 & 24.669 \\
\hline
Average & 25.23 & 22.444 & 21.56 & 20.111 & 26.389 & 23.147 \\
\bottomrule[1pt]
\end{tabular}
}
\vspace{2mm} \\
\resizebox{0.98\linewidth}{!}{%
\begin{tabular}[t]{lcccccc}
\toprule[1pt]
 & \multicolumn{5}{c}{\textbf{Technicolor - SSIM}} \\
\midrule
 & Fabien & Theater & Train & Birthday & Painter & Average \\
\hline
cam01 & 0.846 & 0.665 & 0.71 & 0.69 & 0.789 & 0.74 \\
cam02 & 0.801 & 0.609 & 0.638 & 0.634 & 0.803 & 0.697 \\
cam03 & 0.75 & 0.556 & 0.644 & 0.651 & 0.86 & 0.692 \\
cam04 & 0.835 & 0.697 & 0.726 & 0.715 & 0.823 & 0.759 \\
cam05 & 0.842 & 0.755 & 0.762 & 0.709 & 0.824 & 0.779 \\
cam06 & 0.821 & 0.675 & 0.658 & 0.682 & 0.784 & 0.724 \\
cam07 & 0.776 & 0.62 & 0.619 & 0.675 & 0.761 & 0.69 \\
cam08 & 0.81 & 0.725 & 0.741 & 0.683 & 0.787 & 0.749 \\
cam10 & 0.825 & 0.701 & 0.539 & 0.687 & 0.797 & 0.71 \\
cam11 & 0.82 & 0.687 & 0.665 & 0.726 & 0.826 & 0.745 \\
cam12 & 0.778 & 0.689 & 0.562 & 0.64 & 0.761 & 0.686 \\
cam13 & 0.83 & 0.781 & 0.657 & 0.669 & 0.819 & 0.751 \\
cam14 & 0.85 & 0.753 & 0.612 & 0.713 & 0.79 & 0.744 \\
\hline
Average & 0.814 & 0.686 & 0.656 & 0.683 & 0.802 & 0.728 \\
\bottomrule[1pt]
\end{tabular}
}
\vspace{2mm} \\
\resizebox{0.98\linewidth}{!}{%
\begin{tabular}[t]{lcccccc}
\toprule[1pt]
 & \multicolumn{5}{c}{\textbf{Technicolor - LPIPS}} \\
\midrule
 & Fabien & Theater & Train & Birthday & Painter & Average \\
\hline\addlinespace[0.5ex]
cam01 & 0.305 & 0.355 & 0.24 & 0.245 & 0.218 & 0.273 \\
cam02 & 0.355 & 0.431 & 0.301 & 0.303 & 0.231 & 0.324 \\
cam03 & 0.394 & 0.484 & 0.328 & 0.319 & 0.218 & 0.349 \\
cam04 & 0.34 & 0.342 & 0.225 & 0.24 & 0.218 & 0.273 \\
cam05 & 0.323 & 0.329 & 0.214 & 0.253 & 0.221 & 0.268 \\
cam06 & 0.333 & 0.399 & 0.27 & 0.275 & 0.235 & 0.302 \\
cam07 & 0.367 & 0.447 & 0.307 & 0.291 & 0.266 & 0.336 \\
cam08 & 0.357 & 0.343 & 0.211 & 0.281 & 0.249 & 0.288 \\
cam10 & 0.307 & 0.36 & 0.266 & 0.259 & 0.24 & 0.287 \\
cam11 & 0.304 & 0.38 & 0.249 & 0.235 & 0.225 & 0.278 \\
cam12 & 0.404 & 0.387 & 0.315 & 0.323 & 0.266 & 0.339 \\
cam13 & 0.336 & 0.316 & 0.258 & 0.295 & 0.247 & 0.29 \\
cam14 & 0.287 & 0.347 & 0.255 & 0.241 & 0.238 & 0.274 \\
\hline\addlinespace[0.5ex]
Average & 0.34 & 0.379 & 0.264 & 0.274 & 0.236 & 0.299 \\
\bottomrule[1pt]
\end{tabular}
}
\label{tab:supp-techni-ours}
\end{table}

\begin{table}[H]
\centering
\caption{\textbf{PSNR, SSIM, and LPIPS of our framework on Neural 3D Video.} Three cameras (cam01, cam05 and cam10) are training views.}
\resizebox{0.98\linewidth}{!}{%
\begin{tabular}[t]{lccccccc}
\toprule[2pt]
 & \multicolumn{6}{c}{\textbf{Neural 3D Video - PSNR}} \\
\midrule
 & coffee martini & cook spinach & cut roasted beef & flame salmon & flame steak & sear steak & Average \\
\hline
cam00 & 21.654 & 28.114 & 26.901 & 22.323 & 27.558 & 27.744 & 25.715 \\
cam02 & 25.236 & 25.926 & 25.206 & 24.866 & 25.472 & 25.365 & 25.345 \\
cam03 & - & 25.194 & 24.231 & - & 25.489 & 25.891 & 25.201 \\
cam04 & 25.477 & 31.417 & - & 25.741 & 31.034 & 30.641 & 28.862 \\
cam06 & 20.31 & 27.966 & 26.145 & 21.048 & 27.624 & 27.609 & 25.117 \\
cam07 & 19.602 & 26.638 & 24.25 & 20.449 & 25.344 & 25.759 & 23.674 \\
cam08 & 19.134 & 26.957 & 24.089 & 19.889 & 26.177 & 26.647 & 23.816 \\
cam09 & 20.957 & 26.997 & 26.072 & 21.31 & 27.161 & 27.446 & 24.99 \\
cam11 & 14.669 & 20.251 & 19.145 & 15.398 & 19.277 & 19.349 & 18.015 \\
cam12 & 13.623 & 19.353 & 18.495 & 14.841 & 18.082 & 18.604 & 17.166 \\
cam13 & 15.547 & 20.289 & 20.071 & 16.839 & 19.371 & 20.014 & 18.688 \\
cam14 & 16.705 & 22.012 & 22.252 & 17.309 & 20.462 & 21.098 & 19.973 \\
cam15 & - & 22.182 & 21.68 & 16.824 & 20.618 & 21.2 & 20.501 \\
cam16 & 15.457 & 21.483 & 20.783 & 16.689 & 20.086 & 20.699 & 19.199 \\
cam17 & - & 22.11 & 20.975 & - & 20.894 & 21.646 & 21.406 \\
cam18 & 14.762 & 21.959 & 20.944 & 15.755 & 20.997 & 21.37 & 19.298 \\
cam19 & 17.115 & 22.524 & 21.444 & 17.245 & 22.972 & 23.031 & 20.722 \\
cam20 & 15.736 & 19.319 & 18.569 & 16.248 & 20.715 & 20.014 & 18.434 \\
\hline
Average & 18.399 & 23.927 & 22.426 & 18.923 & 23.296 & 23.563 & 21.905 \\
\bottomrule[2pt]
\end{tabular}
}
\vspace{-1mm} \\
\resizebox{0.98\linewidth}{!}{%
\begin{tabular}[t]{lccccccc}
\toprule[2pt]
 & \multicolumn{6}{c}{\textbf{Neural 3D Video - SSIM}} \\
\midrule
 & coffee martini & cook spinach & cut roasted beef & flame salmon & flame steak & sear steak & Average \\
\hline
cam00 & 0.832 & 0.907 & 0.897 & 0.841 & 0.918 & 0.915 & 0.885 \\
cam02 & 0.875 & 0.917 & 0.906 & 0.87 & 0.907 & 0.905 & 0.897 \\
cam03 & - & 0.919 & 0.893 & - & 0.903 & 0.911 & 0.906 \\
cam04 & 0.889 & 0.941 & - & 0.884 & 0.945 & 0.94 & 0.92 \\
cam06 & 0.767 & 0.898 & 0.862 & 0.774 & 0.899 & 0.894 & 0.849 \\
cam07 & 0.735 & 0.876 & 0.814 & 0.754 & 0.859 & 0.861 & 0.816 \\
cam08 & 0.735 & 0.874 & 0.8 & 0.754 & 0.874 & 0.87 & 0.818 \\
cam09 & 0.786 & 0.892 & 0.868 & 0.8 & 0.893 & 0.892 & 0.855 \\
cam11 & 0.673 & 0.751 & 0.712 & 0.643 & 0.735 & 0.723 & 0.706 \\
cam12 & 0.65 & 0.717 & 0.684 & 0.627 & 0.689 & 0.69 & 0.676 \\
cam13 & 0.67 & 0.772 & 0.76 & 0.676 & 0.756 & 0.753 & 0.731 \\
cam14 & 0.726 & 0.815 & 0.804 & 0.719 & 0.792 & 0.787 & 0.774 \\
cam15 & - & 0.797 & 0.772 & 0.691 & 0.779 & 0.774 & 0.762 \\
cam16 & 0.669 & 0.772 & 0.74 & 0.676 & 0.744 & 0.746 & 0.724 \\
cam17 & - & 0.772 & 0.728 & - & 0.741 & 0.752 & 0.748 \\
cam18 & 0.613 & 0.767 & 0.716 & 0.629 & 0.739 & 0.732 & 0.699 \\
cam19 & 0.667 & 0.785 & 0.758 & 0.656 & 0.79 & 0.792 & 0.741 \\
cam20 & 0.646 & 0.795 & 0.757 & 0.62 & 0.799 & 0.787 & 0.734 \\
\hline
Average & 0.729 & 0.832 & 0.792 & 0.726 & 0.82 & 0.818 & 0.789 \\
\bottomrule[2pt]
\end{tabular}
}
\vspace{-1mm} \\
\resizebox{0.98\linewidth}{!}{%
\begin{tabular}[t]{lccccccc}
\toprule[2pt]
 & \multicolumn{6}{c}{\textbf{Neural 3D Video - LPIPS}} \\
\midrule
 & coffee martini & cook spinach & cut roasted beef & flame salmon & flame steak & sear steak & Average \\
\hline
cam00 & 0.21 & 0.173 & 0.185 & 0.195 & 0.167 & 0.17 & 0.183 \\
cam02 & 0.184 & 0.16 & 0.173 & 0.179 & 0.169 & 0.17 & 0.172 \\
cam03 & - & 0.169 & 0.186 & - & 0.179 & 0.177 & 0.178 \\
cam04 & 0.181 & 0.155 & - & 0.17 & 0.151 & 0.151 & 0.161 \\
cam06 & 0.25 & 0.183 & 0.207 & 0.235 & 0.179 & 0.182 & 0.206 \\
cam07 & 0.273 & 0.183 & 0.223 & 0.252 & 0.191 & 0.192 & 0.219 \\
cam08 & 0.266 & 0.182 & 0.222 & 0.249 & 0.178 & 0.18 & 0.213 \\
cam09 & 0.219 & 0.15 & 0.168 & 0.204 & 0.146 & 0.147 & 0.172 \\
cam11 & 0.37 & 0.317 & 0.339 & 0.372 & 0.326 & 0.329 & 0.342 \\
cam12 & 0.388 & 0.324 & 0.346 & 0.387 & 0.344 & 0.343 & 0.355 \\
cam13 & 0.358 & 0.281 & 0.291 & 0.347 & 0.286 & 0.292 & 0.309 \\
cam14 & 0.313 & 0.252 & 0.261 & 0.312 & 0.259 & 0.264 & 0.277 \\
cam15 & - & 0.265 & 0.285 & 0.327 & 0.271 & 0.276 & 0.284 \\
cam16 & 0.357 & 0.278 & 0.304 & 0.345 & 0.294 & 0.293 & 0.312 \\
cam17 & - & 0.274 & 0.308 & - & 0.291 & 0.288 & 0.29 \\
cam18 & 0.384 & 0.278 & 0.311 & 0.372 & 0.291 & 0.296 & 0.322 \\
cam19 & 0.352 & 0.277 & 0.294 & 0.348 & 0.268 & 0.27 & 0.301 \\
cam20 & 0.373 & 0.279 & 0.307 & 0.367 & 0.27 & 0.272 & 0.311 \\
\hline
Average & 0.298 & 0.232 & 0.259 & 0.291 & 0.237 & 0.238 & 0.258 \\
\bottomrule[2pt]
\end{tabular}
}
\end{table}

\begin{table}[H]
\centering
\caption{\textbf{PSNR, SSIM, and LPIPS of our framework on Nvidia Dynamic Scenes.} Three cameras (cam01, cam06 and cam10) are training views.}
\resizebox{0.98\linewidth}{!}{%
\begin{tabular}[t]{lccccccc}
\toprule[1pt]
 & \multicolumn{6}{c}{\textbf{Nvidia Dynamic Scenes - PSNR}} \\
\midrule
 & jumping & balloon1 & balloon3 & playground & skating & umbrella & Average \\
\hline\addlinespace[0.5ex]
cam02 & 25.867 & 26.076 & 28.318 & 19.736 & 32.098 & 26.946 & 26.507 \\
cam03 & 23.858 & 25.564 & 27.681 & 19.201 & 26.698 & 27.409 & 25.069 \\
cam04 & 25.41 & 25.337 & 26.352 & 18.441 & 30.907 & 27.033 & 25.58 \\
cam05 & 26.698 & 26.248 & 27.455 & 20.261 & 32.017 & 28.36 & 26.84 \\
cam07 & 24.477 & 25.839 & 27.519 & 19.161 & 29.468 & 22.384 & 24.808 \\
cam08 & 25.006 & 24.963 & 21.572 & 19.643 & 30.531 & 22.493 & 24.035 \\
cam09 & 27.679 & 27.486 & 23.516 & 19.941 & 33.348 & 23.373 & 25.89 \\
cam11 & 20.699 & 25.297 & 24.761 & 20.925 & 21.651 & 25.361 & 23.116 \\
cam12 & 19.521 & 20.987 & 20.808 & 19.436 & 21.827 & 25.385 & 21.328 \\
\hline\addlinespace[0.5ex]
Average & 24.357 & 25.311 & 25.331 & 19.638 & 28.727 & 25.416 & 24.807 \\
\bottomrule[1pt]
\end{tabular}
}
\vspace{2mm} \\
\resizebox{0.98\linewidth}{!}{%
\begin{tabular}[t]{lccccccc}
\toprule[1pt]
 & \multicolumn{6}{c}{\textbf{Nvidia Dynamic Scenes - SSIM}} \\
\midrule
 & jumping & balloon1 & balloon3 & playground & skating & umbrella & Average \\
\hline\addlinespace[0.5ex]
cam02 & 0.818 & 0.848 & 0.878 & 0.695 & 0.93 & 0.822 & 0.832 \\
cam03 & 0.756 & 0.82 & 0.88 & 0.626 & 0.826 & 0.818 & 0.788 \\
cam04 & 0.802 & 0.814 & 0.845 & 0.617 & 0.908 & 0.807 & 0.799 \\
cam05 & 0.858 & 0.845 & 0.832 & 0.699 & 0.934 & 0.84 & 0.835 \\
cam07 & 0.78 & 0.83 & 0.848 & 0.625 & 0.919 & 0.758 & 0.793 \\
cam08 & 0.814 & 0.795 & 0.519 & 0.67 & 0.931 & 0.686 & 0.736 \\
cam09 & 0.875 & 0.87 & 0.81 & 0.69 & 0.949 & 0.8 & 0.832 \\
cam11 & 0.784 & 0.836 & 0.805 & 0.74 & 0.894 & 0.813 & 0.812 \\
cam12 & 0.65 & 0.635 & 0.699 & 0.684 & 0.865 & 0.799 & 0.722 \\
\hline\addlinespace[0.5ex]
Average & 0.793 & 0.81 & 0.791 & 0.672 & 0.906 & 0.794 & 0.794 \\
\bottomrule[1pt]
\end{tabular}
}
\vspace{2mm} \\
\resizebox{0.98\linewidth}{!}{%
\begin{tabular}[t]{lccccccc}
\toprule[1pt]
 & \multicolumn{6}{c}{\textbf{Nvidia Dynamic Scenes - LPIPS}} \\
\midrule
 & jumping & balloon1 & balloon3 & playground & skating & umbrella & Average \\
\hline\addlinespace[0.5ex]
cam02 & 0.144 & 0.108 & 0.106 & 0.182 & 0.096 & 0.122 & 0.126 \\
cam03 & 0.197 & 0.12 & 0.108 & 0.187 & 0.156 & 0.147 & 0.152 \\
cam04 & 0.165 & 0.128 & 0.119 & 0.23 & 0.113 & 0.139 & 0.149 \\
cam05 & 0.127 & 0.115 & 0.115 & 0.177 & 0.094 & 0.129 & 0.126 \\
cam07 & 0.176 & 0.126 & 0.12 & 0.219 & 0.13 & 0.178 & 0.158 \\
cam08 & 0.159 & 0.134 & 0.216 & 0.197 & 0.113 & 0.175 & 0.166 \\
cam09 & 0.128 & 0.106 & 0.153 & 0.176 & 0.097 & 0.161 & 0.137 \\
cam11 & 0.185 & 0.126 & 0.143 & 0.151 & 0.138 & 0.137 & 0.147 \\
cam12 & 0.248 & 0.182 & 0.195 & 0.192 & 0.152 & 0.148 & 0.186 \\
\hline\addlinespace[0.5ex]
Average & 0.17 & 0.127 & 0.142 & 0.19 & 0.121 & 0.148 & 0.15 \\
\bottomrule[1pt]
\end{tabular}
}
\label{tab:supp-nvidia-ours}
\end{table}

\end{document}